\documentclass{article}

\usepackage{microtype}
\usepackage{graphicx}
\usepackage{booktabs} %

\usepackage{hyperref}

\usepackage[accepted]{icml2024}

\usepackage{amsmath}
\usepackage{amssymb}
\usepackage{mathtools}
\usepackage{amsthm}

\usepackage[capitalize,noabbrev]{cleveref}

\theoremstyle{plain}

\theoremstyle{definition}

\theoremstyle{remark}

\usepackage[textsize=tiny, textwidth=1.5in, colorinlistoftodos, disable]{todonotes}

\usepackage{tikz}
\usetikzlibrary{positioning, matrix, fit, arrows}

\usepackage{pgfplots}
\pgfplotsset{compat=1.17}
\usepackage{pgfplotstable}
\usepgfplotslibrary{statistics}
\usepgfplotslibrary{patchplots}
\usepgfplotslibrary{fillbetween}
\usepackage{csvsimple}
\usepackage{subcaption}

\usepackage{enumitem}

\pgfplotsset{
    select coords between index/.style 2 args={
        x filter/.code={
            \ifnum\coordindex<#1\fi
            \ifnum\coordindex>#2\fi
        }
    },
    rshift/.style={
        xshift=\pgfkeysvalueof{/pgfplots/rshift scale}
    },
    lshift/.style={
        xshift=-\pgfkeysvalueof{/pgfplots/lshift scale}
    },
    rshift2/.style={
        xshift=\pgfkeysvalueof{/pgfplots/rshift2 scale}
    },
    lshift2/.style={
        xshift=-\pgfkeysvalueof{/pgfplots/lshift2 scale}
    },
    rshift3/.style={
        xshift=\pgfkeysvalueof{/pgfplots/rshift3 scale}
    },
    lshift3/.style={
        xshift=-\pgfkeysvalueof{/pgfplots/lshift3 scale}
    },
    rshift4/.style={
        xshift=\pgfkeysvalueof{/pgfplots/rshift4 scale}
    },
    lshift4/.style={
        xshift=-\pgfkeysvalueof{/pgfplots/lshift4 scale}
    },
    rshift2 scale/.initial=2em,
    rshift scale/.initial=1em,
    rshift3 scale/.initial=0.3em,
    lshift3 scale/.initial=0.3em,
    rshift4 scale/.initial=0.6em,
    lshift4 scale/.initial=0.6em,
    lshift scale/.initial=1em,
    lshift2 scale/.initial=2em,
}

\newcommand{\argmax}[0]{\ensuremath{\operatorname{argmax}}}
\newcommand{\argmin}[0]{\ensuremath{\operatorname{argmin}}}

\newif\ifExtendedVersion
\ExtendedVersiontrue

\icmltitlerunning{Multi-Agent Reinforcement Learning with Hierarchical Coordination for Emergency Responder Stationing}

\begin{document}

\twocolumn[
\icmltitle{Multi-Agent Reinforcement Learning with Hierarchical Coordination for\\Emergency Responder Stationing}

\icmlsetsymbol{equal}{*}

\begin{icmlauthorlist}
\icmlauthor{Amutheezan Sivagnanam}{1}
\icmlauthor{Ava Pettet}{2}
\icmlauthor{Hunter Lee}{2}
\icmlauthor{Ayan Mukhopadhyay}{2}
\icmlauthor{Abhishek Dubey}{2}
\icmlauthor{Aron Laszka}{1}
\end{icmlauthorlist}

\icmlaffiliation{1}{Pennsylvania State University, University Park, PA, USA}
\icmlaffiliation{2}{Vanderbilt University, Nashville, TN, USA}

\icmlcorrespondingauthor{Amutheezan Sivagnanam}{aqs7319@psu.edu}

\icmlkeywords{Deep Reinforcement Learning, Multi-Agent Learning, Spatial Decomposition, Emergency Response, Markov Decision Process}

\vskip 0.3in
]

\printAffiliationsAndNotice{}  %

\begin{abstract}
An emergency responder management (ERM) system dispatches responders, such as ambulances, when it receives requests for medical aid. ERM systems can also proactively reposition responders between predesignated waiting locations to cover any gaps that arise due to the prior dispatch of responders or significant changes in the distribution of anticipated requests.
Optimal repositioning is computationally challenging due to the exponential number of ways to allocate responders between locations and the uncertainty in future requests. 
The state-of-the-art approach in proactive repositioning is a hierarchical approach based on spatial decomposition and online Monte Carlo tree search, which may require minutes of computation for each decision in a domain where seconds can save lives.
We address the issue of long decision times by introducing a novel reinforcement learning (RL) approach, based on the same hierarchical decomposition, but replacing online search with learning.
To address the computational challenges posed by large, variable-dimensional, and discrete state and action spaces, we propose: (1) actor-critic based agents that incorporate transformers to handle variable-dimensional states and actions, (2) projections to fixed-dimensional observations to handle complex states, and (3) combinatorial techniques to map continuous actions to discrete allocations.
We evaluate our approach using real-world data from two U.S. cities, Nashville, TN and Seattle, WA. 
Our experiments show that 
compared to the state of the art, our approach reduces computation time per decision by three orders of magnitude, while also slightly reducing average ambulance response time by 5 seconds.
\end{abstract}

\section{Introduction}

Dynamically repositioning resources under uncertainty is an important problem in societal-scale cyber-physical systems~\cite{pettet2021hierarchical2}, such as bike repositioning~\cite{li2018dynamic}, ride-hailing and public transit~\cite{jin2019coride,xi2021ddrl,talusan2024online}, and emergency response management (ERM)~\cite{mukhopadhyay2020review}.
In such problems, a decision-maker must sequentially optimize the allocation of resources in space and time to respond to uncertain demand (e.g., calls for service).
We focus specifically on ERM, a critical problem faced by urban communities across the globe. Indeed, 240 million emergency medical service calls are made in the U.S. alone each year~\cite{mukhopadhyay2020review}.
Whenever a request for medical aid is reported to an ERM, a responder is dispatched to the scene of the incident.
Responders administer critical services on the scene (e.g., %
basic life support), transfer the patient to the nearest hospital, and head back to their assigned waiting locations (called \textit{depots}) to wait until their next dispatch. Due to the critical nature of emergency medical aid, dispatching decisions are typically constrained to greedy policies that send the nearest available responder~\cite{mukhopadhyay2020review}.
However, it is possible to proactively optimize the %
waiting locations of the responders
so that expected response times for future incidents are minimized~\cite{pettet2021hierarchical2,pettet2020algorithmic}. 
 
The problem of proactive %
repositioning
is computationally challenging due to the uncertainty in future demand and the combinatorial state-action space of the problem---the number of possible responder assignments grows exponentially with the number of responders. For example, in one of our experimental settings, %
Nashville, TN,
the number of possible allocations at each decision epoch is on the order of $10^{33}$. 
Prior works use \textit{centralized}, \textit{decentralized}, and \textit{hierarchical} approaches to solve the ERM reallocation problem. %
\cite{ji2019deep} propose a learning-based approach for centralized %
reallocation,
which can reallocate a responder each time it %
finishes serving a request. Unfortunately, such centralized solutions with a monolithic state-action space~\cite{mukhopadhyay2018decision,ji2019deep} do not scale to large ERM systems.  Decentralized approaches~\cite{pettet2020algorithmic}, on the other hand, split the state-action space such that each responder makes its own decisions, sacrificing coordination between the responders to achieve scalability, which can lead to sub-optimal decisions.

The state-of-the-art approach~\cite{pettet2021hierarchical2} applies hierarchical planning to the responder allocation problem, which in principle lies midway between the centralized and decentralized approaches. This approach first partitions the spatial area under consideration into smaller and more manageable areas called \textit{regions}.
A \textit{low-level planner} uses Monte-Carlo tree search (MCTS), an online technique, to solve each region-specific problem independently. 
A \textit{high-level planner} allocates responders between the regions based on expected demand. \cite{pettet2021hierarchical2}'s hierarchical approach alleviates scalability concerns while maintaining coordination between nearby responders,
and \textit{comprehensively outperforms other methods in this domain}. However, the framework’s use of MCTS for low-level planning requires running extensive simulations at decision time, which need significant computation time for decision-making (between 2--4 minutes). This is infeasible in practice since delays in reallocation during a coverage gap (when an area does not have any available responders) could be catastrophic in ERM. 
This raises an interesting research problem: \textit{How can we reduce decision time to avoid delays in making reallocation decisions without sacrificing solution quality compared to the current state of the art?}

We address this problem by introducing a novel learning-based approach that can make reallocation decisions in a fraction of a second. However, applying a learning-based approach directly to the problem formulation of \cite{pettet2021hierarchical2} is challenging
since: \textbf{(1)} even with the spatial decomposition, both \emph{state and action spaces are high-dimensional and discrete}; \textbf{(2)} low-level planners must be able to handle \emph{variable-dimensional state and action spaces} as the number of responders in a region may vary over time; and \textbf{(3)} \emph{rewards are very noisy}, especially for the high-level planner, since response times can vary widely depending on the locations and times of incidents.
We tackle these challenges systematically in this paper.

Specifically, we make the following contributions.
\textbf{(1)} We introduce a multi-agent reinforcement-learning with hierarchial coordination for responder repositioning by replacing the high- and low-level planners of \cite{pettet2021hierarchical2}'s framework with \emph{actor-critic based agents that can handle complex, high-dimensional action spaces}. 
\textbf{(2)} To handle complex, high-dimensional states, we introduce \textit{projections from states to low-dimensional features}, which capture relevant state information.
\textbf{(3)} We incorporate \textit{transformers into low-level agents to handle variable-dimensional states and actions}.
\textbf{(4)} To facilitate gradient-based actor training, we introduce \textit{efficient combinatorial optimizations that take continuous actions from an actor and map them to discrete allocations}.
\textbf{(5)} To reduce noise in the high-level agent's rewards, we \textit{estimate its rewards using the low-level agents' critics}. 
\textbf{(6)}
We evaluate our approach using real-world %
data from Nashville and Seattle, two cities in the U.S., and show that our approach not only \emph{reduces the computational time per decision by multiple orders of magnitude} but also slightly reduces ambulance response times compared to the state of the art. %

\newcommand{\AllIncidents}[0]{\mathcal{I}}
\newcommand{\Events}[0]{\mathcal{Y}_{\textit{t}}}
\newcommand{\Incidents}[0]{\mathcal{I}_{\textit{t}}}
\newcommand{\Reallocations}[0]{\mathcal{RA}_{\textit{t}}}
\newcommand{\RegionEvents}[0]{{\Events}^{\OneRegion}}
\newcommand{\RegionIncidents}[0]{{\Incidents}^{\OneRegion}}
\newcommand{\RegionReallocations}[0]{{\Reallocations}^{\OneRegion}}

\newcommand{\Cells}[0]{\mathcal{C}}
\newcommand{\Regions}[0]{\mathcal{G}}
\newcommand{\TravelModel}[0]{\mathcal{M}}
\newcommand{\RegionCells}[0]{\mathcal{C}^{\OneRegion}}
\newcommand{\Depots}[0]{\mathcal{D}}
\newcommand{\Responders}[0]{\mathcal{V}}
\newcommand{\Assignments}[0]{\mathcal{A}}
\newcommand{\Hospitals}[0]{\mathcal{H}}
\newcommand{\RegionDepots}[0]{\Depots^{\OneRegion}}
\newcommand{\RegionResponders}[0]{\Responders^{\OneRegion}}
\newcommand{\ResponderPositions}[0]{\mathcal{P}_{\textit{t}}}

\newcommand{\SingleResponderPosition}[0]{\mathcal{P}_{\textit{t}}^{\textit{v}}}
\newcommand{\OneRegion}[0]{\textit{g}}
\newcommand{\SingleRegion}[0]{\OneRegion \in \Regions}
\newcommand{\SingleDepot}[0]{\textit{d} \in \Depots^{\OneRegion}}
\newcommand{\SingleResponder}[0]{\textit{v} \in \Responders^{\OneRegion}}
\newcommand{\SingleCell}[0]{\textit{c} \in \Cells}
\newcommand{\DepotIncidentRate}[0]{\lambda^d_{\textit{t}}}
\newcommand{\RegionIncidentRate}[0]{\lambda^\OneRegion_{\textit{t}}}
\newcommand{\CellIncidentRate}[0]{\lambda^c_{\textit{t}}}
\newcommand{\DepotOccupancy}[0]{\eta[\textit{d}]}
\newcommand{\AverageArrivalTime}[0]{\beta[\textit{d}]}
\newcommand{\IncidentRate}[0]{\lambda}
\newcommand{\ActionValue}[0]{\alpha_{\{\textit{v},\textit{d}\}}}
\newcommand{\ExpectedArrivalTime}[0]{\phi_t[\textit{d},\textit{v}]}
\newcommand{\HLPAction}[0]{\textbf{a}_{\textit{t}}^{\textit{h}}}
\newcommand{\NormalizedHLPAction}[0]{\textbf{a}_{\textit{t}}^{\textit{H}}}
\newcommand{\LLPAction}[0]{\textbf{a}_{\textit{t}}^{\OneRegion}}
\newcommand{\SingleLLPAction}[0]{\LLPAction[\textit{d},\textit{v}]}
\newcommand{\TransHLPAction}[0]{\textbf{A}_{\textit{t}}}
\newcommand{\TransPrevHLPAction}[0]{\textbf{A}_{\textit{t} - 1}}

\newcommand{\TransLLPAction}[0]{\textbf{A}_{\textit{t}}^{\OneRegion}}
\newcommand{\LLPActionHat}[0]{\textbf{a}_{\textit{t}}^{\hat{\OneRegion}}}
\newcommand{\SumLLPAction}[0]{\sum_{\OneRegion}^{\Regions} \LLPAction}
\newcommand{\SumHatLLPAction}[0]{\sum_{\hat{\OneRegion}}^{\Regions} \LLPActionHat}
\newcommand{\SumTransLLPAction}[0]{\sum_{\OneRegion}^{\Regions} \TransLLPAction}
\newcommand{\SumTransHatLLPAction}[0]{\sum_{\hat{\OneRegion}}^{\Regions} \TransLLPActionHat}
\newcommand{\SumHLPAction}[0]{\sum_{\OneRegion}^{\Regions} \HLPAction}
\newcommand{\SumHatHLPAction}[0]{\sum_{\hat{\OneRegion}}^{\Regions} \HLPAction}
\newcommand{\SumHatNormalizedHLPAction}[0]{\sum_{\hat{\OneRegion}}^{\Regions} \NormalizedHLPAction}
\newcommand{\SumTransHLPAction}[0]{\sum_{\OneRegion}^{\Regions} \TransHLPAction}
\newcommand{\SumTransHatHLPAction}[0]{\sum_{\hat{\OneRegion}}^{\Regions} \TransLLPActionHat}

\newcommand{\Nodes}[0]{\mathcal{Y}}
\newcommand{\SingleNode}[0]{y \in \Nodes}
\newcommand{\SSNodes}[0]{\textbf{Y}}

\newcommand{\Edges}[0]{\mathcal{E}}

\newcommand{\DepotCapacity}[0]{\textbf{DC}}
\newcommand{\DepotOccupants}[0]{\textbf{DO}}
\newcommand{\NearestHospital}[0]{\textbf{H}}

\newcommand{\State}[0]{s_t}
\newcommand{\RegionState}[0]{s^\OneRegion_t}
\newcommand{\Action}[0]{a_t}
\newcommand{\RegionAction}[0]{a^\OneRegion_t}
\newcommand{\RegionResponderPositions}[0]{\mathcal{P}^{\OneRegion}_{\textit{t}}}

\begin{figure*}
    \centering
    \resizebox{\textwidth}{!}{\input{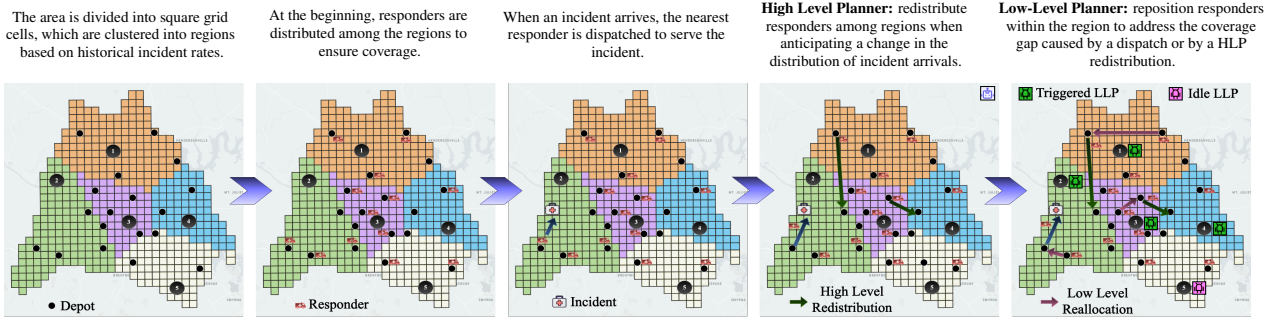}}
    \caption{High-level overview of state-of-the-art hierarchical framework~\cite{pettet2021hierarchical2}, described in \cref{subsec:hierarchical_framework}.}
    \label{fig:high-level-architecture}
\end{figure*}

\section{Problem Formulation}
\label{sec:problem}

We begin by introducing the assumptions of the ERM reallocation problem and %
modeling it %
as a continuous-time Markov decision process. %
Then, in \cref{subsec:hierarchical_framework}, we describe the hierarchical decision-making framework that enables tractable decision-making. 
This problem formulation and  hierarchical framework are  based on state-of-the-art prior work~\cite{pettet2021hierarchical2,mukhopadhyay2020review}. We introduce our novel learning-based %
framework in \cref{sec:solution}.

\subsection{Model}
\label{subsec:model}
An ERM system manages a set of resonders to serve requests\footnote{We use \textit{requests} and \textit{incidents} synonymously.} for medical aid that are distributed over space and time and are unknown in advance. 
The ERM's spatial area of operation is divided into a grid of equally sized square cells ($\Cells$). Requests for medical aid follow a spatio-temporal probability distribution that can be modeled using independent Poisson distributions for each cell. For a specific cell $c \in \Cells$, we denote the expected rate of incident occurrence at time $t$ for some duration (e.g., over the next hour) by $\CellIncidentRate$.  

The ERM system consists of the following components: a set of responders ($\Responders$) to serve requests, a fixed set of spatially-located depots ($\Depots$) where responders idle between serving requests, and a fixed set of hospitals ($\Hospitals$) where patients are taken after being picked up at a request's location. When a request is reported, the ERM system assigns the nearest available responder to service it (this is governed by the practical constraints that ERM operators face). If no responder is available, the incident enters a waiting queue. Once a responder arrives at the request's location, it treats the patient on-scene for a fixed time ($t_{\textit{serve}}$),  after which it transports the patient to the nearest hospital. The movement of responders between cells follows a time-varying travel model ($\TravelModel$):  the time taken to move from cell $c_i \in \Cells$ to cell $c_j \in \Cells$ at time $t$ is given by $\TravelModel(c_i, c_j, t)$. After servicing a request, the responder is either immediately dispatched to the request at the top of the waiting queue or returns to an empty depot to wait if the queue is empty. 

We model the ERM's stochastic decision-making problem of dynamically reallocating responders ($\Responders$) to depots ($\Depots$) in anticipation of future incidents as a continuous-time Markov decision process (MDP). The objective of the MDP is to reduce the system's expected incident response time (i.e., cumulative time taken to reach the scene of each incident after it is reported). We describe the MDP below.   

\paragraph{State} At time $t$, the system state is denoted by $s_t$ and consists of the set of incidents waiting for service $\Incidents$, the incident rates $\CellIncidentRate, \forall{\SingleCell}$ of the cells, and the state $\ResponderPositions$ of the responders.  For each responder $v \in \Responders$, the state %
$p^{\textit{v}}_{\textit{t}} \in \ResponderPositions$ is a tuple $\langle d^{\textit{v}}$, $i^{\textit{v}}$, $h^{\textit{v}}$, $c^{\textit{v}}$, $t^v_{\textit{avail}}\rangle$, where $d^{\textit{v}} \in \Cells$ is the depot where responder $v$ is assigned to wait when it is not serving an incident; $i^{\textit{v}} \in \Cells$ is the location of the incident to which responder $v$ is currently assigned; $h^{\textit{v}} \in \Cells$ is the location of the hospital to which responder $v$ is currently taking a patient; $c^{\textit{v}} \in \Cells$ is responder $v$'s current location; and $t^v_{\textit{avail}} \in \mathbb{R}^{+}$ is the point in time at which responder $v$ will become available, i.e., when it will drop off the patient at the assigned hospital.  
Variables $i^{\textit{v}},h^{\textit{v}},$ and $t^v_{\textit{avail}}$ are empty if responder $v$ is not assigned to an incident at time $t$.

\paragraph{Transition}
The environment has two types of events: incident occurrences and incident rate changes. We define two types of decision epochs based on these events: (1) when a responder is dispatched to serve an incident and (2) when changes in incident rates are detected. Between two decision epochs, the state of the system evolves as responders serve their incidents and move to their assigned depots according to the travel model $\TravelModel$.
We assume the assignment of responders to depots does not change between consecutive decision epochs since no new information becomes available.

\paragraph{Action} 
An action is the reallocation of some responders $v \in \Responders$ from their currently assigned depot $d^v_t$ to a different depot $d^v_{t+1}$ ($d^v_t \neq d^v_{t+1}$) at decision epoch $t$.

\paragraph{Reward}
If the next decision epoch is due to the arrival of an incident, then the reward for the last action is the response time for the incident.
On the other hand, if the next decision epoch is due to a change in the incident rates, in which case there is no incident to serve, the reward is 0.
Note that in the ERM setting, the goal is to find an allocation that minimizes the expected incident response times. Therefore, despite using the standard term ``reward,'' the objective is to minimize the expected discounted rewards.

\subsection{Hierarchical Decision Framework} \label{subsec:hierarchical_framework}

A key challenge to solving the MDP defined in \cref{subsec:model} is the combinatorial nature of the state and action spaces. We address this scalability challenge by using the hierarchical decision-making framework %
introduced by \cite{pettet2021hierarchical2}. We assume that the ERM's spatial area has been divided into a set of smaller, more manageable regions $\Regions$ based on the historical incident rates (using the same approach as \cite{pettet2021hierarchical2}), with $\RegionCells$ denoting the cells assigned to region $\SingleRegion$.
Decision-making for these regions is decomposed into two stages: high-level and low-level decision making. First, the high-level decision agent  distributes the available responders between the regions. We denote the region to which responder $v$ is assigned as $\OneRegion^v \in \Regions$. Then, low-level agents optimize the responder assignments independently within each region. This significantly reduces the complexity of each region's assignment subproblem, as the low-level agents need to consider only the interactions between responders and depots within a single region.

In this hierarchical decision-making framework, %
the following general procedure is followed each time an incident is reported: (1) The nearest available responder is dispatched. (2) The high-level planner decides if the allocation  of responders to regions is unbalanced, and if so, actuates an appropriate redistribution of responders among the regions. (3) The low-level planner is invoked for every region if the high-level planner changed the region distribution; otherwise, it is invoked only for the region from which the responder was dispatched (in the first step) to address any coverage gaps that arose due to the dispatch. \cref{fig:high-level-architecture} shows a high-level overview of the hierarchical framework.

\section{Solution Approach}
\label{sec:solution}

Now, we introduce our novel learning-based %
approach for proactive repositioning in ERM.
We utilize the hierarchical framework from the state-of-the-art approach~\cite{pettet2021hierarchical2}, but we replace both the \textit{high-level planner} (HLP) and the \textit{low-level planners} (LLPs) with deep reinforcement learning agents to overcome the long decision-making time of online search. Due to the critical nature of ERM, any additional time spent on planning can have a negative impact on serving future incidents \cite{jaldell2014time}. 
However, the application of reinforcement learning faces several computational obstacles.
Even after the hierarchical decomposition, action spaces remain vast, which poses challenges for finding optimal actions.
We propose actor-critic-based agents for both the HLP and the LLPs, specifically, the DDPG algorithm~\cite{lillicrap2015continuous}, since a trained actor can choose an action at a very low computational cost. In \ifExtendedVersion 
\cref{app:why_ddpg}\else 
Appendix G in \cite{sivagnanam2024multiagent}\fi, we explain the rationale behind  our choice of applying  DDPG in more detail.
However, this leads to another challenge since such agents are ill-suited for discrete action spaces; we address this by letting actors choose continuous actions, which we map to discrete allocations using efficient combinatorial optimization: minimum-cost flow for HLP and maximum-weight matching for LLPs.

Similarly, state spaces remain vast even after decomposition; hence, we map states to low-dimensional feature vectors, which capture relevant information, for both LLPs and the HLP.
LLPs also face the challenge of variable-dimensional state and action spaces due to the varying number of responders in a region; we tackle this by incorporating transformers into the actors.
Finally, the HLP faces the challenge of noisy rewards that are weakly correlated to its actions since each response time depends on only one of many regions (i.e., the region where the incident occurred%
); we address this by estimating HLP rewards using LLP critics.

\subsection{Low-Level Decision Agent: Reallocating Responders within a Region}
\label{sec:low_level}

\begin{figure*}[ht!]
    \centering
     \resizebox{\textwidth}{!}
    {\input{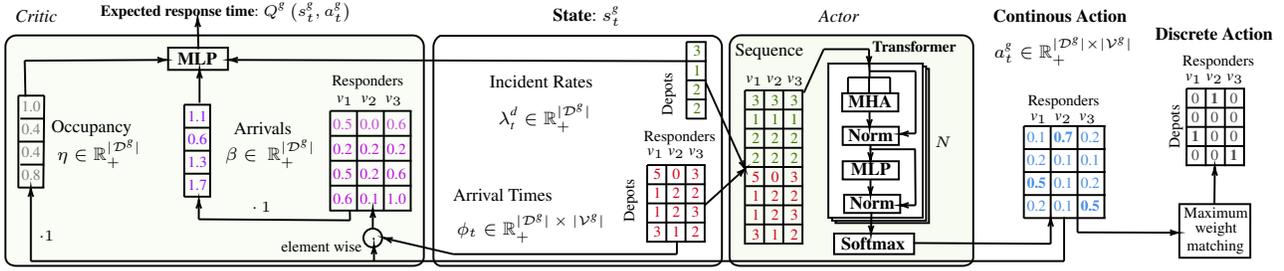}}
    \caption{Overview of the low-level RL agent training process using DDPG for a region $\SingleRegion$. First, we map the complex, variable-dimensional state ($s^\textit{g}_{\textit{t}}$) to a sequence of feature vectors, which we feed to the \textit{actor} to obtain a continuous action ($\LLPAction$). Next, we discretize the continuous action using \textit{maximum weight matching} to allocate responders within the region $\OneRegion$. Finally, we use the \textit{critic} to judge the performance of the \textit{actor} by feeding the state and action as fixed-sized vectors to the \textit{critic} and perform learning against response time to serve the incident.}
    \label{fig:low_level_process}
\end{figure*}

We first introduce an MDP formulation of the problem of repositioning responders \textit{within} a region by a low-level planner. Then, we explain how we build on the DDPG algorithm~\cite{lillicrap2015continuous} to train an agent. %
\cref{fig:low_level_process} provides an overview of the architecture of our low-level~agent. 

We formulate each region's reallocation problem as an MDP as follows:
\textbf{State:} The state  $\RegionState$ of region $g$ consists of the incident rates of the cells in region $\OneRegion$ ($\CellIncidentRate, \forall{c \in \RegionCells}$) and  the states of responders currently assigned to $\OneRegion$ ($\RegionResponderPositions = \{p^v_t \mid p^v_t \in \ResponderPositions \land \SingleResponder\}$). \textbf{Transition:} 
We consider two decision epochs: (1) when a responder assigned to region $\OneRegion$ is dispatched to a request, and (2) when the set of responders assigned to $\OneRegion$ is changed by the high-level agent (described in detail in \cref{sec:high_level}). \textbf{Action:} An action $\Assignments[\OneRegion]$ is a repositioning of responders between depots in the region~$\OneRegion$. \textbf{Reward:} We consider the same reward as the original MDP, i.e., response time, but calculated only for incidents served by the responders in region $g$.

\paragraph{Actor Input} 
To implement low-level planning, 
we apply actor-critic based RL to the MDP above.
To tackle the challenge of high-dimensional state space, we transform the region state $\RegionState$ into the following two sets of features:

\begin{itemize}[topsep=0pt]
    \item \textit{Arrival time} \(\ExpectedArrivalTime\): total time that each responder \(\SingleResponder\) would take to reach each depot \(\SingleDepot\) if responder $\SingleResponder$ were assigned to idle at depot \(d\) after completing its current task. We let \(\ExpectedArrivalTime=0\) if responder \(v\) is currently at depot \(d\).  Intuitively,  \(\ExpectedArrivalTime\) captures how soon responder \(v\) would be ready to serve incidents nearby depot \(d\). Arrival time \(\ExpectedArrivalTime\) is computed as:
\[
    \ExpectedArrivalTime = 
    \begin{cases}
    \TravelModel(c^v, d, t) & \text{if } t^v_{\textit{\tiny{avail}}} < t\\
    t^v_{\textit{\tiny{avail}}} - t + \TravelModel(h^v, d, t^v_{\textit{\tiny{avail}}})              & \text{otherwise.}
    \end{cases}
\]

    \item \textit{Nearby incident rate} $\DepotIncidentRate$:  sum of the incident rates~$\CellIncidentRate$ at time~$t$ for cells $c\in \RegionCells$ that are \textit{near} depot $d$. Specifically, $
    \lambda^d_t = \sum_{c \;\in\; \textbf{NearCells}(d, t)} \lambda^c_t
    $,
    where $
    \textbf{NearCells}(d, t)$ is the set of cells for which $d$ is the closest depot, and is computed as $ \{{c} | {c} \in \Cells, \argmin_{\hat{d} \in \Depots} \TravelModel(c, \hat{d}, t) = d \}$. Intuitively, $\DepotIncidentRate$ estimates the future demand for which depot $d$ is likely to be ``responsible.''
\end{itemize}

\paragraph{Actor Network}
We describe the key elements of the actor network here;
see \ifExtendedVersion 
\cref{app:trxl_actor} \else 
Appendix B.1 in \cite{sivagnanam2024multiagent} \fi
for a more detailed description.
We feed a sequence  of  feature vectors $\langle \ExpectedArrivalTime, \DepotIncidentRate \;|\; \forall{\SingleDepot} \rangle$, one feature vector for each responder ${\SingleResponder}$, into the actor network to obtain a sequence of likelihood vectors $\LLPAction[v]$, one likelihood vector for each responder ${\SingleResponder}$. 
The actor network is based on Transformer-XL \cite{dai2019transformer}, consisting of 
$N$ sequential TrXL layers, which enable ``coordination'' between responders based on their input features. %
We apply \textit{softmax} activation after the TrXL layers to output %
likelihood values $\LLPAction[v]$, where $\LLPAction[v] \cdot 1 = 1, \;\LLPAction[v] \geq 0$, which assign responder $v$ to depots in the region. 
We combine the likelihood vectors of all the responders in the region to obtain the continuous action~$\LLPAction$. 

\paragraph{Discrete Action}
Since the output $\LLPAction$ of the actor network is continuous, it is necessary to discretize actions for actual allocation.
We compute a discrete assignment of responders to depots ($\Assignments[\OneRegion]$) by finding a maximum weight matching \cite{duan2014linear,kuhn1955hungarian} for matrix $\LLPAction$, which maximizes the linear sum of likelihood values for responders and their assigned depots (see \ifExtendedVersion 
\cref{app:discrete_low_level} \else 
Appendix B.2 in \cite{sivagnanam2024multiagent} \fi for more details). This enables us to efficiently compute a discrete assignment that is most similar to the continuous actor output.

\paragraph{Critic} 
To handle complex state and action spaces, the critic relies on the following three sets of features, which are computed from $s_t$ and  $\LLPAction$ for each depot $\SingleDepot$:

\begin{itemize}[topsep=0pt]
    \item \textit{Depot occupancy} $\DepotOccupancy$: overall likelihood of some responder being assigned to depot $d$. Occupancy $\DepotOccupancy$ is computed by summing the corresponding likelihood values in $\LLPAction$ and truncating the sum to be between 0 and 1, i.e., $
 \DepotOccupancy = \text{Clip}(\sum_{v}^{\RegionResponders} \SingleLLPAction, 0, 1) $. 
Intuitively, $\DepotOccupancy$ is the heuristic chance that at least one responder is assigned to depot $\SingleDepot$.

    \item \textit{Likely available time} $\AverageArrivalTime$: weighted sum of the arrival times of responders to depot $d$. Available time $\AverageArrivalTime$ is computed as: $
\AverageArrivalTime = \sum_{v}^{\RegionResponders} \ExpectedArrivalTime \cdot \SingleLLPAction \,\forall{\SingleDepot}
$.
    Intuitively, in combination with $\DepotOccupancy$, time $\AverageArrivalTime$ indicates how soon a responder is expected to arrive at the depot. %

    \item \textit{Nearby incident rate} $\DepotIncidentRate$: same as $\DepotIncidentRate$ in actor input. %
\end{itemize}

We feed these feature vectors into the critic network, which is a multi-layer perceptron, to obtain the estimated average response time $Q^\OneRegion_t(s^\OneRegion_t, a^\OneRegion_t)$ for action $a^\OneRegion_t$ in state $s^\OneRegion_t$:
\[
    Q^\OneRegion_t(s^\OneRegion_t, a^\OneRegion_t) = \textbf{MLP}(\langle  \DepotOccupancy, \AverageArrivalTime, \DepotIncidentRate \;\;|\;\; \forall{\SingleDepot}\rangle)
\]

\begin{figure*}[ht!]
    \centering
    \resizebox{\textwidth}{!}{\input{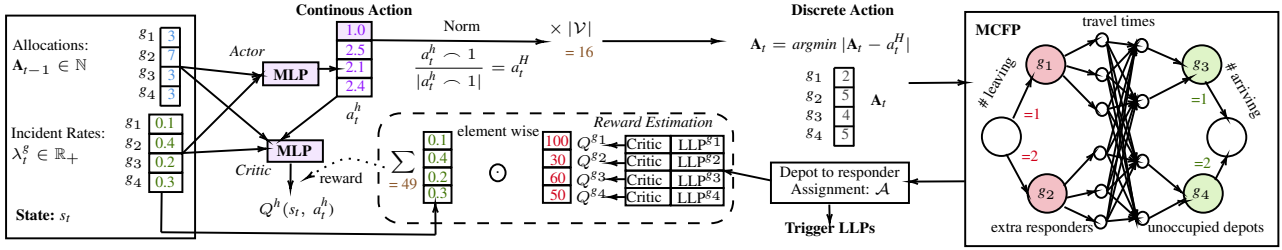}}
    \caption{Overview of the training process of the high-level RL agent using DDPG. First, we map the state to a fixed-size feature vector and feed it into the MLP-based \textit{actor} to generate the \textit{continuous action} ($\HLPAction$). Next, we discretize the continuous action and feed it into the \textit{minimum-cost flow problem} to generate the assignment of responders to depots ($\mathcal{A}$). After that, we trigger those LLPs whose regions were affected by the high-level reallocation. Finally, we use the \textit{critic} to judge the performance of the \textit{actor} by training the critic with \emph{rewards estimated by the LLP critics}.}
    \label{fig:high_level_process}
\end{figure*}

\subsection{High-Level Decision Agent: Reallocating Responders between Regions}
\label{sec:high_level}

We first introduce an MDP formulation of the problem of reallocating responders \textit{among} regions by a high-level planner. %
Then, we explain how we apply DDPG, %
mapping states to feature vectors, %
discretizing actions, and estimating HLP rewards from LLP critics. \cref{fig:high_level_process} provides an overview of the architecture of the high-level agent. %

We formulate the high-level redistribution problem as an MDP as follows: \textbf{State:} We consider the same state representation as the original MDP ($\State$), which consists of the incident rates of all cells  ($\CellIncidentRate, \forall{c \in \Cells}$) and the state of all responders ($\ResponderPositions$). \textbf{Transition:} Transitions occur whenever  incident rates change. During a transition, responders are redistributed among the regions based on the actions described below.  \textbf{Action:} An action specifies which responders to move from one region to another.

To find optimal actions given the MDP for the high-level planner, we use a similar approach as the low-level planner. First, we leverage an abstraction of input features that represent the complex state space. Then, we use DDPG to train the high-level using our novel reward estimation. We discuss the reward estimation in detail later in this section.

\paragraph{Actor}
We overcome the complexity of the state space by mapping the state to the following features for each region $\SingleRegion$:

\begin{itemize}[topsep=0pt]
    \item \textit{Region incident rates} $\RegionIncidentRate$: sum of incident rates for all cells in region $\SingleRegion$ at time $\textit{t}$ ($\RegionIncidentRate = \sum_c^{\RegionCells} \CellIncidentRate$).
    \item \textit{Allocation} $\TransPrevHLPAction[\OneRegion]$: number of responder currently assigned to the region $\SingleRegion$.
\end{itemize}

\paragraph{Actor Network} We feed the input feature vectors into the actor network, which is a multi-layer perceptron, to obtain the action $\HLPAction$, which is a non-negative continuous vector of dimension $|\Regions| - 1$:
\[
     \HLPAction = \textbf{MLP}(\langle \lambda^\OneRegion_t, \TransPrevHLPAction[\OneRegion] \mid \forall{\SingleRegion} \rangle)
\]

We let the actor network compute values for first $|\Regions| - 1$ regions that provides the ratio of responders with respect to the last region. We obtain the distribution of responders to regions ($\NormalizedHLPAction$) by appending a constant value of 1 to the vector $\HLPAction$ and perform normalization $\NormalizedHLPAction =\frac{\HLPAction \frown 1}{|\HLPAction \frown 1|}$. 

\paragraph{Discrete Action}
Based on the normalized action $\NormalizedHLPAction$, we finally compute the number of responders $\TransHLPAction$ assigned to each region $g$, where $\TransHLPAction = \argmin |\TransHLPAction - \NormalizedHLPAction|$, satisfying the following constraints: number of responders assigned to any region $\SingleRegion$ should not exceeds the number of depots available in the region ($\TransHLPAction[\OneRegion] \leq |\RegionDepots|,\;\forall{\SingleRegion}$), and all the responders must be assigned to one region ($\sum_{\OneRegion}^{\Regions}\TransHLPAction[\OneRegion] = |\Responders|$). Since this computation is trivial, we provide a detailed description in the technical appendix (see \ifExtendedVersion 
\cref{app:redistribution}\else 
Appendix C.1 in \cite{sivagnanam2024multiagent}\fi).

\paragraph{Assignment of Responders to Depots} 
Once we discretize the action and obtain the number of responders allocated to each region, we must choose the specific responders to reallocate and which depots they should be assigned to in their new regions. We tackle this problem by transforming it into a standard minimum-cost flow problem (MCFP) \cite{ahuja1993network}, and solve it by minimizing the total expected travel time for the responders to reach their newly assigned depots (for details, see \ifExtendedVersion 
\cref{app:high_level_assignment}\else 
Appendix C.2 in \cite{sivagnanam2024multiagent}\fi). Thereby, we obtain the assignment $\mathcal{A}$, which allocates responders $\Responders_{\textit{leaves}}$ to depots $\Depots_{\textit{unoccupied}}$.

\paragraph{Reward Estimation} 
It is non-trivial to estimate the rewards for the high-level planner -- the redistribution of responders is  \textit{rewarded} only when responders are dispatched to incidents efficiently, which in turn, depends on the allocation chosen by the low level planner. Hence, the rewards from the regions are noisy. Further, events happen in regions without synchronization, and each LLP transitions independently. To tackle these issues, we introduce a novel reward estimation technique for the high-level RL agent: we use the low-level RL agents' critics to estimate the value of each region's allocation. We hypothesize that these low-level critic values estimate how well a specific distribution of responders across regions helps reduce response times.  In addition, since incident rates can vary across decision-making epochs, we also incorporate each region's incident rate to prioritize regions based on forecasted demand. Accordingly, we estimate the reward for the high-level RL agent as a weighted sum of critic values ($\sum_\OneRegion^{\Regions} \RegionIncidentRate \cdot Q_\OneRegion^{*} (s^\OneRegion_{\textit{t}}, \LLPAction | \theta^{Q_\OneRegion^{*}})$) from the low-level agents corresponding to each region $\SingleRegion$.

\paragraph{Critic} We feed the feature vectors for the critic network (i.e., the feature vectors for the actor network to represent the state + action) and compute the value ($Q_t(s_t, \HLPAction)$) of performing the action $\HLPAction$ at the  state $s_t$ as follows:
\[
    Q_t(s_t, \HLPAction) = \textbf{MLP}(\langle \lambda^\OneRegion_t, \TransPrevHLPAction[\OneRegion] \mid \forall{\SingleRegion}, \HLPAction  \rangle)
\]

\definecolor{ColorCustomGreen}{rgb}{0, 0.8, 0}
\colorlet{ColorOur}{ColorCustomGreen}
\colorlet{ColorLegendOur}{ColorOur!50}
\colorlet{ColorLegendOurThick}{ColorOur!99}
\colorlet{ColorMCTS}{red}
\colorlet{ColorLegendMCTS}{ColorMCTS!50}
\colorlet{ColorLegendMCTSThick}{ColorMCTS!99}
\colorlet{ColorStatic}{purple}
\colorlet{ColorLegendStatic}{ColorStatic!50}
\colorlet{ColorDynamic}{brown}
\colorlet{ColorLegendDynamic}{ColorDynamic!50}
\colorlet{ColorPMedian}{teal}
\colorlet{ColorLegendPMedian}{ColorPMedian!50}
\colorlet{ColorPMedianHalf}{olive}
\colorlet{ColorLegendPMedianHalf}{ColorPMedianHalf!50}
\colorlet{ColorPMedianFull}{lime}
\colorlet{ColorLegendPMedianFull}{ColorPMedianFull!50}
\colorlet{ColorVPG}{orange}
\colorlet{ColorLegendVPG}{ColorVPG!50}
\colorlet{ColorOurMMC}{blue}
\colorlet{ColorLegendOurMMC}{ColorOurMMC!50}
\colorlet{ColorGTransMMC}{magenta}
\colorlet{ColorLegendGTransMMC}{ColorGTransMMC!50}
\colorlet{ColorRNNMMC}{black}
\colorlet{ColorLegendRNNMMC}{ColorRNNMMC!50}
\colorlet{ColorRandom}{gray}
\colorlet{ColorLegendRandom}{ColorRandom!50}
\colorlet{ColorLegendRandomThick}{ColorRandom!99}

\section{Numerical Results}
\label{sec:numerical}

\begin{figure*}
\begin{subfigure}{0.235\textwidth}
\pgfplotstableread[col sep=comma,]{data_nashville/city_level_summary_5_24.csv}\regionfive
\pgfplotstableread[col sep=comma,]{data_nashville/city_level_summary_6_24.csv}\regionsix
\pgfplotstableread[col sep=comma,]{data_nashville/city_level_summary_7_24.csv}\regionseven
\begin{tikzpicture}
\begin{axis}[
      boxplot/draw direction=y,
      width=\linewidth,
      xtick={1,2,3},
      xticklabels={{5}, {6}, {7}},
      xmin=0.5,
      xmax=3.5,
      height = 4.0cm,
      ymajorgrids,
      major grid style={draw=gray!25},
      bugsResolvedStyle/.style={},
      xlabel={Num. of Regions},
      font=\small,
    ]

\addplot+[boxplot={box extend=0.10, draw position=1}, ColorOur, solid, lshift4, fill=ColorOur!20, mark=x] table [col sep=comma, y=drl] {\regionfive};
\addplot+[boxplot={box extend=0.10, draw position=1}, ColorMCTS, solid, lshift3, fill=ColorMCTS!20, mark=x] table [col sep=comma, y=mcts] {\regionfive};
\addplot+[boxplot={box extend=0.10, draw position=1}, ColorPMedian, solid, fill=ColorPMedian!20, mark=x] table [col sep=comma, y=pmedian10] {\regionfive};
\addplot+[boxplot={box extend=0.10, draw position=1}, ColorDynamic, solid, rshift3, fill=ColorDynamic!20, mark=x] table [col sep=comma, y=greedy] {\regionfive};
\addplot+[boxplot={box extend=0.10, draw position=1}, ColorStatic, solid, rshift4, fill=ColorStatic!20, mark=x] table [col sep=comma, y=static] {\regionfive};

\addplot+[boxplot={box extend=0.10, draw position=2}, ColorOur, solid, lshift4, fill=ColorOur!20, mark=x] table [col sep=comma, y=drl] {\regionsix};
\addplot+[boxplot={box extend=0.10, draw position=2}, ColorMCTS, solid, lshift3, fill=ColorMCTS!20, mark=x] table [col sep=comma, y=mcts] {\regionsix};
\addplot+[boxplot={box extend=0.10, draw position=2}, ColorPMedian, solid, fill=ColorPMedian!20, mark=x] table [col sep=comma, y=pmedian10] {\regionsix};
\addplot+[boxplot={box extend=0.10, draw position=2}, ColorDynamic, solid, rshift3, fill=ColorDynamic!20, mark=x] table [col sep=comma, y=greedy] {\regionsix};
\addplot+[boxplot={box extend=0.10, draw position=2}, ColorStatic, solid, rshift4, fill=ColorStatic!20, mark=x] table [col sep=comma, y=static] {\regionsix};

\addplot+[boxplot={box extend=0.10, draw position=3}, ColorOur, solid, lshift4, fill=ColorOur!20, mark=x] table [col sep=comma, y=drl] {\regionseven};
\addplot+[boxplot={box extend=0.10, draw position=3}, ColorMCTS, solid, lshift3, fill=ColorMCTS!20, mark=x] table [col sep=comma, y=mcts] {\regionseven};
\addplot+[boxplot={box extend=0.10, draw position=3}, ColorPMedian, solid, fill=ColorPMedian!20, mark=x] table [col sep=comma, y=pmedian10] {\regionseven};
\addplot+[boxplot={box extend=0.10, draw position=3}, ColorDynamic, solid, rshift3, fill=ColorDynamic!20, mark=x] table [col sep=comma, y=greedy] {\regionseven};
\addplot+[boxplot={box extend=0.10, draw position=3}, ColorStatic, solid, rshift4, fill=ColorStatic!20, mark=x] table [col sep=comma, y=static] {\regionseven};

\end{axis}
\end{tikzpicture}
\caption{24 Responders}
\label{fig:city_level_response_times_twenties_nashville_twenty_four}
\end{subfigure}
\hspace{0.01em}
\begin{subfigure}{0.235\textwidth}
\pgfplotstableread[col sep=comma,]{data_nashville/city_level_summary_5_26.csv}\regionfive
\pgfplotstableread[col sep=comma,]{data_nashville/city_level_summary_6_26.csv}\regionsix
\pgfplotstableread[col sep=comma,]{data_nashville/city_level_summary_7_26.csv}\regionseven

\begin{tikzpicture}
\begin{axis}[
      boxplot/draw direction=y,
            width=\linewidth,
      xtick={1,2,3},
      xticklabels={{5}, {6}, {7}},
      xmin=0.5,
      xmax=3.5,
      height = 4.0cm,
      ymajorgrids,
      major grid style={draw=gray!25},
      bugsResolvedStyle/.style={},
      xlabel={Num. of Regions},
      font=\small,
    ]

\addplot+[boxplot={box extend=0.10, draw position=1}, ColorOur, solid, lshift4, fill=ColorOur!20, mark=x] table [col sep=comma, y=drl] {\regionfive};
\addplot+[boxplot={box extend=0.10, draw position=1}, ColorMCTS, solid, lshift3, fill=ColorMCTS!20, mark=x] table [col sep=comma, y=mcts] {\regionfive};
\addplot+[boxplot={box extend=0.10, draw position=1}, ColorPMedian, solid, fill=ColorPMedian!20, mark=x] table [col sep=comma, y=pmedian10] {\regionfive};
\addplot+[boxplot={box extend=0.10, draw position=1}, ColorDynamic, solid, rshift3, fill=ColorDynamic!20, mark=x] table [col sep=comma, y=greedy] {\regionfive};
\addplot+[boxplot={box extend=0.10, draw position=1}, ColorStatic, solid, rshift4, fill=ColorStatic!20, mark=x] table [col sep=comma, y=static] {\regionfive};

\addplot+[boxplot={box extend=0.10, draw position=2}, ColorOur, solid, lshift4, fill=ColorOur!20, mark=x] table [col sep=comma, y=drl] {\regionsix};
\addplot+[boxplot={box extend=0.10, draw position=2}, ColorMCTS, solid, lshift3, fill=ColorMCTS!20, mark=x] table [col sep=comma, y=mcts] {\regionsix};
\addplot+[boxplot={box extend=0.10, draw position=2}, ColorPMedian, solid, fill=ColorPMedian!20, mark=x] table [col sep=comma, y=pmedian10] {\regionsix};
\addplot+[boxplot={box extend=0.10, draw position=2}, ColorDynamic, solid, rshift3, fill=ColorDynamic!20, mark=x] table [col sep=comma, y=greedy] {\regionsix};
\addplot+[boxplot={box extend=0.10, draw position=2}, ColorStatic, solid, rshift4, fill=ColorStatic!20, mark=x] table [col sep=comma, y=static] {\regionsix};

\addplot+[boxplot={box extend=0.10, draw position=3}, ColorOur, solid, lshift4, fill=ColorOur!20, mark=x] table [col sep=comma, y=drl] {\regionseven};
\addplot+[boxplot={box extend=0.10, draw position=3}, ColorMCTS, solid, lshift3, fill=ColorMCTS!20, mark=x] table [col sep=comma, y=mcts] {\regionseven};
\addplot+[boxplot={box extend=0.10, draw position=3}, ColorPMedian, solid, fill=ColorPMedian!20, mark=x] table [col sep=comma, y=pmedian10] {\regionseven};
\addplot+[boxplot={box extend=0.10, draw position=3}, ColorDynamic, solid, rshift3, fill=ColorDynamic!20, mark=x] table [col sep=comma, y=greedy] {\regionseven};
\addplot+[boxplot={box extend=0.10, draw position=3}, ColorStatic, solid, rshift4, fill=ColorStatic!20, mark=x] table [col sep=comma, y=static] {\regionseven};

\end{axis}
\end{tikzpicture}
\caption{26 Responders}
\label{fig:city_level_response_times_twenties_nashville_twenty_six}
\end{subfigure}
\hspace{0.01em}
\begin{subfigure}{0.235\textwidth}
\pgfplotstableread[col sep=comma,]{data_nashville/city_level_summary_5_28.csv}\regionfive
\pgfplotstableread[col sep=comma,]{data_nashville/city_level_summary_6_28.csv}\regionsix
\pgfplotstableread[col sep=comma,]{data_nashville/city_level_summary_7_28.csv}\regionseven

\begin{tikzpicture}
\begin{axis}[
      boxplot/draw direction=y,
        width=\linewidth,
      xtick={1,2,3},
      xticklabels={{5}, {6}, {7}},
      xmin=0.5,
      xmax=3.5,
      height = 4.0cm,
      ymajorgrids,
      major grid style={draw=gray!25},
      bugsResolvedStyle/.style={},
      xlabel={Num. of Regions},
      font=\small,
    ]

\addplot+[boxplot={box extend=0.10, draw position=1}, ColorOur, solid, lshift4, fill=ColorOur!20, mark=x] table [col sep=comma, y=drl] {\regionfive};
\addplot+[boxplot={box extend=0.10, draw position=1}, ColorMCTS, solid, lshift3, fill=ColorMCTS!20, mark=x] table [col sep=comma, y=mcts] {\regionfive};
\addplot+[boxplot={box extend=0.10, draw position=1}, ColorPMedian, solid, fill=ColorPMedian!20, mark=x] table [col sep=comma, y=pmedian10] {\regionfive};
\addplot+[boxplot={box extend=0.10, draw position=1}, ColorDynamic, solid, rshift3, fill=ColorDynamic!20, mark=x] table [col sep=comma, y=greedy] {\regionfive};
\addplot+[boxplot={box extend=0.10, draw position=1}, ColorStatic, solid, rshift4, fill=ColorStatic!20, mark=x] table [col sep=comma, y=static] {\regionfive};

\addplot+[boxplot={box extend=0.10, draw position=2}, ColorOur, solid, lshift4, fill=ColorOur!20, mark=x] table [col sep=comma, y=drl] {\regionsix};
\addplot+[boxplot={box extend=0.10, draw position=2}, ColorMCTS, solid, lshift3, fill=ColorMCTS!20, mark=x] table [col sep=comma, y=mcts] {\regionsix};
\addplot+[boxplot={box extend=0.10, draw position=2}, ColorPMedian, solid, fill=ColorPMedian!20, mark=x] table [col sep=comma, y=pmedian10] {\regionsix};
\addplot+[boxplot={box extend=0.10, draw position=2}, ColorDynamic, solid, rshift3, fill=ColorDynamic!20, mark=x] table [col sep=comma, y=greedy] {\regionsix};
\addplot+[boxplot={box extend=0.10, draw position=2}, ColorStatic, solid, rshift4, fill=ColorStatic!20, mark=x] table [col sep=comma, y=static] {\regionsix};

\addplot+[boxplot={box extend=0.10, draw position=3}, ColorOur, solid, lshift4, fill=ColorOur!20, mark=x] table [col sep=comma, y=drl] {\regionseven};
\addplot+[boxplot={box extend=0.10, draw position=3}, ColorMCTS, solid, lshift3, fill=ColorMCTS!20, mark=x] table [col sep=comma, y=mcts] {\regionseven};
\addplot+[boxplot={box extend=0.10, draw position=3}, ColorPMedian, solid, fill=ColorPMedian!20, mark=x] table [col sep=comma, y=pmedian10] {\regionseven};
\addplot+[boxplot={box extend=0.10, draw position=3}, ColorDynamic, solid, rshift3, fill=ColorDynamic!20, mark=x] table [col sep=comma, y=greedy] {\regionseven};
\addplot+[boxplot={box extend=0.10, draw position=3}, ColorStatic, solid, rshift4, fill=ColorStatic!20, mark=x] table [col sep=comma, y=static] {\regionseven};

\end{axis}
\end{tikzpicture}
\caption{28 Responders}
\label{fig:city_level_response_times_twenties_nashville_twenty_eight}
\end{subfigure}
\hspace{0.01em}
\begin{subfigure}{0.27\textwidth}
\pgfplotstableread[col sep=comma,]{data_nashville/city_level_summary_5_26.csv}\regionfive
\pgfplotstableread[col sep=comma,]{data_nashville/city_level_summary_6_26.csv}\regionsix
\pgfplotstableread[col sep=comma,]{data_nashville/city_level_summary_7_26.csv}\regionseven

\begin{tikzpicture}
\begin{axis}[
      boxplot/draw direction=y,
      xtick={1,2,3},
      xticklabels={{5}, {6}, {7}},
      xmin=0.5,
      xmax=3.5,
      height = 4.0cm,
      width=\linewidth,
      ymajorgrids,
      major grid style={draw=gray!25},
      bugsResolvedStyle/.style={},
      xlabel={Num. of Regions},
      font=\small,
    ]

\addplot+[boxplot={box extend=0.05, draw position=1}, ColorMCTS, solid, lshift, fill=ColorMCTS!20, mark=x] table [col sep=comma, y=mcts] {\regionfive};
\addplot+[boxplot={box extend=0.05, draw position=1}, ColorOurMMC, solid, lshift3, fill=ColorOurMMC!20, mark=x] table [col sep=comma, y=drl_trans_mmc] {\regionfive};
\addplot+[boxplot={box extend=0.05, draw position=1}, ColorGTransMMC, solid, rshift3, fill=ColorGTransMMC!20, mark=x] table [col sep=comma, y=drl_gtrans_mmc] {\regionfive};
\addplot+[boxplot={box extend=0.05, draw position=1}, ColorRNNMMC, solid,rshift, fill=ColorRNNMMC!20, mark=x] table [col sep=comma, y=drl_rnn_mmc] {\regionfive};

\addplot+[boxplot={box extend=0.05, draw position=2}, ColorMCTS, solid, lshift, fill=ColorMCTS!20, mark=x] table [col sep=comma, y=mcts] {\regionsix};
\addplot+[boxplot={box extend=0.05, draw position=2}, ColorOurMMC, solid, lshift3, fill=ColorOurMMC!20, mark=x] table [col sep=comma, y=drl_trans_mmc] {\regionsix};
\addplot+[boxplot={box extend=0.05, draw position=2}, ColorGTransMMC, solid, rshift3, fill=ColorGTransMMC!20, mark=x] table [col sep=comma, y=drl_gtrans_mmc] {\regionsix};
\addplot+[boxplot={box extend=0.05, draw position=2}, ColorRNNMMC, solid,rshift, fill=ColorRNNMMC!20, mark=x] table [col sep=comma, y=drl_rnn_mmc] {\regionsix};

\addplot+[boxplot={box extend=0.05, draw position=3}, ColorMCTS, solid, lshift, fill=ColorMCTS!20, mark=x] table [col sep=comma, y=mcts] {\regionseven};
\addplot+[boxplot={box extend=0.05, draw position=3}, ColorOurMMC, solid, lshift3, fill=ColorOurMMC!20, mark=x] table [col sep=comma, y=drl_trans_mmc] {\regionseven};
\addplot+[boxplot={box extend=0.05, draw position=3}, ColorGTransMMC, solid, rshift3, fill=ColorGTransMMC!20, mark=x] table [col sep=comma, y=drl_gtrans_mmc] {\regionseven};
\addplot+[boxplot={box extend=0.05, draw position=3}, ColorRNNMMC, solid,rshift, fill=ColorRNNMMC!20, mark=x] table [col sep=comma, y=drl_rnn_mmc] {\regionseven};

\end{axis}
\end{tikzpicture}
\caption{Different Architectures}
\label{fig:city_level_response_times_llp_others_nashville}
\end{subfigure}
\caption{Distribution of average response times (lower is better) with our approach (\textcolor{ColorLegendOur}{$\blacksquare$}), MCTS (\textcolor{ColorLegendMCTS}{$\blacksquare$}),  $p$-median with $\alpha$ = 1.0 (\textcolor{ColorLegendPMedian}{$\blacksquare$}), greedy policy (\textcolor{ColorLegendDynamic}{$\blacksquare$}), and static policy, i.e., no proactive repositioning (\textcolor{ColorLegendStatic}{$\blacksquare$}) for 10 different sample incident chains with (a) 24 responders, (b) 26 responders, and (c) 28 responders. (d) distribution of average response times using MCTS (\textcolor{ColorLegendMCTS}{$\blacksquare$}) and various architectures as the actor for the low-level agent (TrXL (\textcolor{ColorLegendOurMMC}{$\blacksquare$}), GTrXL (\textcolor{ColorLegendGTransMMC}{$\blacksquare$}), and LSTM (\textcolor{ColorLegendRNNMMC}{$\blacksquare$})), trained and evaluated with the HLP from prior work \cite{pettet2021hierarchical2} for 10 different sample incident chains with 26 responders (Nashville).}
\label{fig:city_level_response_times_twenties_nashville}
\end{figure*}
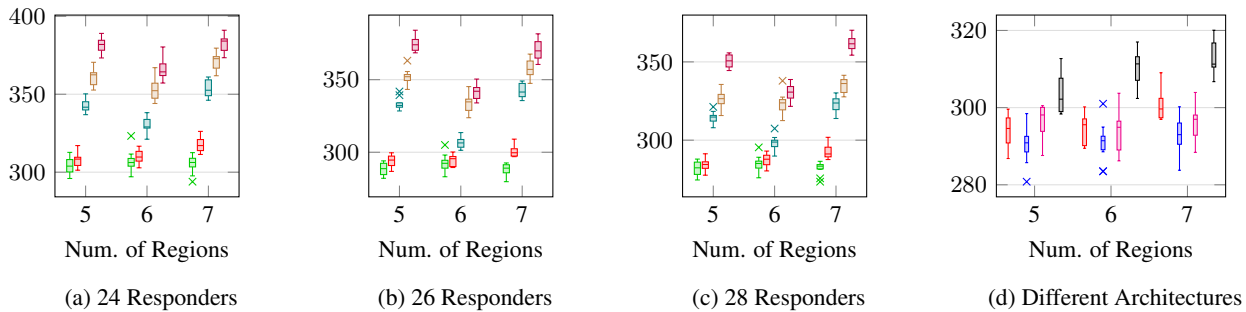

\subsection{Dataset and Experiment Setup}
\label{subsec:dataset}

We evaluate our approach using real-world data from two U.S. cities.  First, we apply our approach to emergency response data from Nashville, TN, which was published by \cite{pettet2021hierarchical2}. This data includes processed incident chains, incident rates, depot locations, hospital locations,  operational data provided by the Nashville fire department, 
and  dynamic travel times (i.e., travel times that vary by time of the day and by day of the week) generated using contraction hierarchies \cite{geisberger2008contraction} and OSRM \cite{huber2016calculate}. A one-by-one mile square grid is applied to the city, which aligns with the configurations followed by local authorities. The city has 36 depots, 9 general hospitals, and 26 responders to assist emergency response management. Further, the data contains 60 incident chains sampled from the historical incident data distribution and corresponding  rate data. Each chain spans 11 days and contains between 1818 and 2025 incidents, averaging 1907 incidents per chain. The data also includes several region segmentations $\Regions$ of the city's cells, obtained by applying the $k$-means clustering algorithm based on historical incident rates in cells $\Cells$ and geographic proximity. We also evaluate our algorithm using publicly available data from another U.S. city, Seattle, WA~\cite{seattleData}. Due to limited space, we present the Seattle results in \ifExtendedVersion 
\cref{app:seatle_numerical}\else 
Appendix D.3 of \cite{sivagnanam2024multiagent}\fi. Our implementation and data are available as part of the supplementary material.

We assume that depots can accommodate at most one responder at a time (i.e., $\DepotCapacity(d) = 1, \forall{d \in \Depots})$, same as \cite{pettet2021hierarchical2}. In practice, we could create extra single-capacity depots to represent depots that can accommodate more than one responder, and apply the same solution approach. We assume that the time taken to serve an incident ($t_{\textit{serve}}$) is constant 20 minutes (same as \cite{pettet2021hierarchical2}). %

\subsection{Baselines}
We compare our approach against the state-of-the-art MCTS approach~\cite{pettet2021hierarchical2} and three other baselines. For the other baselines, we leverage the high-level planner from \cite{pettet2021hierarchical2}. We set the initial state of the environment by distributing the responders among the region based on the high-level planner of \cite{pettet2021hierarchical2}, and sequentially assign responders to depots in each region.
Specifically, we use the following baselines in addition to \cite{pettet2021hierarchical2}: (1) $p$-\textit{median-based policy}: we use a modified $p$-median formulation suggested by \cite{vazirizade2021learning}, which incorporates a \textit{balancing} term to account for account for responders becoming unavailable when attending incidents; (2) \textit{greedy policy}: we make reallocation decisions based on the expected incident rates and expected travel times to reach the depots; and (3) \textit{static policy}: a baseline where the allocation is never changed from the initial one.
Note that the static policy closely resembles real-world strategies followed by first-responders in practice~\cite{pettet2021hierarchical2}. We describe the baselines in the technical appendix (\ifExtendedVersion 
\cref{app:baseline_description}\else 
Appendix D.1 of \cite{sivagnanam2024multiagent}\fi).

For the baseline experiments, %
we trigger both the high- and low-level agents each time an incident occurs or when there have been no incidents in the last 60 minutes, same as suggested in the original work by \cite{pettet2021hierarchical2}. However, when evaluating our approach, we trigger the high-level agent each time the predicted incident rates change, with at least a 60-minute interval after the previous trigger;  and trigger the low-level agent for a region whenever responders enter or leave the region, or serve an incident. For MCTS baseline, we use the same hyperparameters as \cite{pettet2021hierarchical2}: iteration limit is 1000, discount factor is 0.99995, UCT trade-off parameter is 1.44, and number of generated samples is 50.

\subsection{Training}
\label{subsec:training}
We train both the high- and low-level agents using 50 sampled incident chains and evaluate the trained models using the remaining 10 chains. 
We use the Adam optimizer \cite{kingma2014adam} and a learning rate of 10$^{-3}$. We set the reward decay rate 
($\gamma$) for low-level agents to 0.5.
We run all algorithms on an Intel Xeon E5-2680 28-core CPU with 128GB of RAM. We detail the agent-specific training~below.

\paragraph{Low-Level RL Agent}
We train a low-level agent for each region $\OneRegion$ individually using our approach and using the DRLSN baseline \cite{ji2019deep}. We vary the number of responders assigned to the region during training randomly based on a binomial distribution with values ranging from 1 to $|\RegionDepots|$ and with probability $\frac{|\Responders|}{|\Depots|}$. After the arrival of an incident (from the sampled incident chain) or after 1 hour without any incidents, we select a reallocation action using the RL policy based on the current state of the environment (one specific region). We apply the reallocation action to the environment and capture the response time of serving the next incident as a negative reward for the reallocation action. We use a fixed-size experience-replay buffer, where the experiences are stored based on a First-In-First-Out (FIFO) policy. Accordingly, we add the experience tuple (i.e., state, action, reward, next state) for each transition to the buffer. During the learning process, we randomly sample a fixed-size batch of experiences from the buffer, and we train the actor and critic networks using the DDPG algorithm. After each transition step, we update the actor and critic networks until they converge to the optimal RL policy.

We perform an architecture search based on the experiences gathered from online RL to obtain the best hyper-parameters for our approach's TrXL-based actor network (described in detail in \ifExtendedVersion 
\cref{app:architecture_search}\else 
Appendix D.4 of \cite{sivagnanam2024multiagent}\fi). For our critic network, we use a feed-forward network with one hidden layer with 64 neurons and ReLU activation, followed by a dropout layer with a dropout rate of 0.1 and an output layer with one neuron and linear activation. For our baseline DRLSN, we consider an MLP with three hidden layers containing 256, 128, and 64 neurons respectively and use ReLU activation in all the~layers.

\paragraph{High-Level RL Agent}
The high-level agent is trained after completing the low-level agent training.  At the beginning of each episode, we provide an initial allocation of responders to regions that is proportional to the region-level incident rates (i.e., sum of the incident arrival rates of all cells in the region), and the low-level agent of each region is invoked to perform an initial allocation of responders to depots within its region. Whenever the distribution of incident rates changes at the regional level, we select a high-level redistribution action using the RL policy. We apply the redistribution action to the environment, invoke low-level agents as needed, and capture the weighted sum of the low-level critics as a reward for the redistribution action. We use a fixed-size replay buffer with a FIFO policy to store experiences (i.e., tuples of state, action, reward, and next state). During the learning process, we randomly sample batches of experience from the replay buffer, and we train the actor and critic networks using DDPG. After each transition step, we update the actor and critic networks until convergence. We vary the number of responders between $26 \pm 3$ in each episode during training.

For the actor, we use a feed-forward network with two hidden layers of 256 and 64 neurons, respectively, with ReLU activation, followed by a dropout layer with a dropout rate of 0.1 and an output layer with one neuron and linear activation. For the critic, we use a feed-forward network with one hidden layer of 64 neurons with ReLU activation, followed by a dropout layer with a dropout rate of 0.1 and an output layer with one neuron and linear activation.

\paragraph{Training Time}

Our low-level agent using TrXL takes around 1 day ($\approx$ 100 episodes) to converge to an optimal policy for regions with 8 or less depots. For regions with 9 or more depots, it takes up to 14 days after obtaining the best architecture. Our high-level RL agent takes around 2 days ($\approx$ 40 episodes) to converge.

\subsection{Evaluation}
\label{subsec:evaluation}

\paragraph{Computation Times}

On average, our DDPG-based approach takes around 0.22 seconds to make a single decision, considerably less than the MCTS baseline \cite{pettet2021hierarchical2}, which takes 3 minutes. 
While a 3-minute delay might not necessarily seem significant, it is actually substantial in the context of this problem setting: the average travel time between depots is around 15 minutes in Nashville, and the average response time is around 4.8 minutes. In a domain where seconds can save lives, an additional 3-minute delay is very significant. While other baselines may take less time (DRLSN: 0.045 seconds; $p$-median: 0.20 seconds; greedy: 0.10 seconds), we point out that a latency of 0.22 seconds is negligible in practice for ERM.

\paragraph{Response Times}

\cref{fig:city_level_response_times_twenties_nashville_twenty_four,fig:city_level_response_times_twenties_nashville_twenty_six,fig:city_level_response_times_twenties_nashville_twenty_eight} show average response times for our DDPG-based approach compared against baseline approaches (MCTS, $p$-median-based policy, greedy policy, and static policy) using 10 sampled incident chains. On average, we outperform the state-of-the-art MCTS baseline by 5 seconds in 5-region segmentation and by 13 seconds in 7-region segmentation, which are significant savings in the ERM domain~\cite{pettet2021hierarchical2,mayer1979emergency}. We observe that $p$-median, greedy, and static policy always perform poorly compared to the state-of-the-art approach. We include a comparison with the DRLSN baseline in the technical appendix as both MCTS and the proposed approach outperform it by a significant margin (see \ifExtendedVersion 
\cref{app:additional_nashville}\else 
Appendix D.2 in \cite{sivagnanam2024multiagent}\fi). 
In addition, we also train our LLP agent considering the entire city to be a single region, and evaluate it using the same incident chains. We find that this centralized variant performs 66 seconds worse than our hierarchical approach. 

\paragraph{Ablation Study: Low-Level Agents with Different Architectures}

\cref{fig:city_level_response_times_llp_others_nashville} shows the distribution of average response times using different low-level actor architectures---TrXL,  Gated TrXL (GTrXL)~\cite{parisotto2020stabilizing,parisotto2021efficient}, and LSTM---trained and evaluated with the HLP from prior work~\cite{pettet2021hierarchical2}, alongside the MCTS baseline. %
We observe that TrXL-based low-level agents perform better than other architectures. %
TrXL-based low-level agents reduce time to serve an incident by 5 seconds on average compared to GTrXL and by 18 seconds on average compared to LSTM. Accordingly, we train the high-level RL agent with TrXL-based low-level~agents.

\section{Related Work}
\label{sec:related}

We provide a brief discussion of the most closely related prior work here; for a broader discussion, please see \ifExtendedVersion \cref{app:extended_related_work}\else 
Appendix F of \cite{sivagnanam2024multiagent}\fi.

\citet{zhang2021multi} classify Multi-Agent Reinforcement Learning (MARL) based on the type of agents (i.e., homogeneous or heterogeneous), objective (i.e., cooperative, competitive, or combination of both cooperative and competitive \cite{lowe2017multi}), control mechanism (i.e., centralized or decentralized), and learning paradigm (i.e., singular or hierarchical). In our approach, we consider cooperative heterogeneous agents with hierarchical coordination and the shared goal of minimizing response times for future incidents. Most cooperative MARLs maximize a shared reward when a centralized controller controls it or maximize an average reward in a decentralized setting \cite{oroojlooy2023review,zhang2018fully,kar2012qd}. In our approach, we train the low-level agents to maximize rewards independently~\cite{eghtesad2024multi}. Then, we introduce a reward-estimation mechanism for the high-level agent (i.e., convex combination of low-level critic values) to coordinate the independent execution of low-level agents.

In the domain of resource reallocation, \citet{jin2019coride} apply a Multi-Agent Hierarchical Reinforcement Learning (MA-HRL) approach, based on Feudal RL~\cite{dayan1992feudal,vezhnevets2017feudal}. To make each agent aware of other agents at the same hierarchy level, \citet{jin2019coride} introduce attention between the agents; whereas in our approach, we let the low-level agents act independently of each other; instead, we utilize the high-level agent for coordination. %

\section{Conclusion}
We introduce a novel multi-agent RL-based approach %
by replacing the high- and low-level planners of \citeauthor{pettet2021hierarchical2}'s framework with deep reinforcement learning, addressing the computational challenges faced by learning. 
We show using real-world data that 
our approach reduces computation time per decision  by %
\emph{three orders of magnitude}
compared to the state of the art. %
We also confirm the general advantage of hierarchical approaches over centralized ones as a centralized variant of our learning-based approach performs poorly in comparison. Finally, we  demonstrate that redistributing responders when incident rates change is better than redistributing after every incident. %

\section*{Software and Data}

Software code and data are available online \cite{sivagnanam2024multidataset}.
Within the ZIP file \texttt{MARL-HC-ERS-ICML24.zip},
we provide our complete code base as well as implementations for all baselines, i.e., DRLSN baseline \cite{ji2019deep}, MCTS baseline \cite{pettet2021hierarchical2}, $p$-median baseline \cite{vazirizade2021learning}, greedy policy, and static policy. 
For instructions on training and evaluation, please see \texttt{README.txt} (available in the root folder of the extracted ZIP file).

\section*{Acknowledgments}

This material is based upon work sponsored by the National Science Foundation under Grants CNS-1952011 and CNS-2238815 and by the Department of Energy under Award Number DE-EE0009212. The authors acknowledge the computing support from NSF Chameleon.
Any opinions, findings, and conclusions or recommendations expressed in this material are those of the authors and do not necessarily reflect the views of the National Science Foundation or of the Department of Energy.
The authors would like to thank the anonymous reviewers of ICML'24 for their insightful feedback and suggestions.

\section*{Impact Statement}

With our approach, emergency response management systems can make proactive reallocation decisions in a fraction of a second, ensuring better services for all incoming requests. 
Making reallocation decisions in a fraction of a second is extremely beneficial in ERM, a domain where seconds may mean the difference between life and death.
Our approach also slightly reduces the average response time compared to the current state-of-the-art approach, which requires extensive computation for each decision.
Overall, these improvements have the potential to shorten emergency response times, leading to a positive societal impact.
Further, our research did not involve any personally identifiable or sensitive information.

\bibliography{main}
\bibliographystyle{icml2024}

\ifExtendedVersion
\newpage
\appendix
\newpage

\section{Notation}
\label{app:notations}

\cref{tab:symbols} provides a summary of the most important notation used throughout our paper.

\begin{table}[t]
\centering
\renewcommand{\arraystretch}{1.1}
\setlength{\tabcolsep}{4pt}
\caption{List of Symbols}
\label{tab:symbols}
\begin{tabular}{cp{6.5cm}} %
\toprule
\textbf{Symbol} & \multicolumn{1}{c}{\textbf{Description}}  \\ \midrule
\multicolumn{2}{c}{Constants}  \\ \midrule
$\Regions$ & Regions \\ \hline
$\Depots$  & Depots  \\ \hline
$\Hospitals$ & Hospitals \\ \hline
$\Cells$ & Cells \\ \hline
$\TravelModel$ & Travel model  \\ \hline
$\RegionDepots$  & Depots in region $\SingleRegion$  \\ \hline
$\Responders$  & Responders  \\ \midrule
\multicolumn{2}{c}{Variables}  \\ \midrule
$\RegionResponders$  & Responders assigned to region $\SingleRegion$ \\ \hline
$\HLPAction$ & High-level RL agent action  \\ \hline
$\LLPAction$ & Low-level RL agent action \\ \hline
$\TransHLPAction$  & Redistribution of responders to regions \\ \hline
$\TransHLPAction[\OneRegion]$  & Number of responders assigned to region $\SingleRegion$ \\ 
\hline
$\Assignments$ & Assignment of responders to depots  \\ \hline
$\Assignments[\OneRegion]$ & Assignment of responders to depots in region $\SingleRegion$  \\ \hline
$\CellIncidentRate$  & Incident rate in cell $\SingleCell$ at time $t$ \\ \hline
$\Incidents$ & Incidents waiting for service at time $t$ \\ \hline
$\ResponderPositions$  & State of responders at time $t$ \\ \hline
$g^{v}$  & Region assigned to responder $v$  \\ \hline
$d^{v}$  & Depot assigned to responder $v$ \\ \hline
$i^{v}$  & Incident assigned to responder $v$  \\ \hline
$h^{v}$  & Hospital assigned to responder $v$  \\ \hline
$c^{v}$  & Current position of responder $v$ \\ \hline
$t^{v}_{\textit{avail}}$ & Time at which responder $\textit{v}$ will  be available \\ \hline
$\ExpectedArrivalTime$ & Total time responder $\textit{v}$ takes to reach  depot $\textit{d}$ after completing its current task at time $t$ \\ \bottomrule
\end{tabular}
\label{table:symbol_list}
\end{table}

\section{Low-Level Decision Agent}

\subsection{Architecture of TrXL-based Actor}
\label{app:trxl_actor}

We build our actor network with TrXL layers~\cite{dai2019transformer} as its fundamental units. 
A TrXL layer consists of two main components: multi-head attention (MHA) and multilayer perceptron (MLP). 
After each main component is a normalization layer (Norm-MHA and Norm-MLP). Our actor-network contains $N$ layers of TrXL followed by a \textit{Softmax} layer at the end. 
We use $\textit{Input}_X^L$ to denote the input of component $X \in \{$MHA, Norm-MHA, MLP, Norm-MLP$\}$ at  layer $L \in \{1, 2, 3, \ldots, N\}$, and $\textit{Output}_X^L$ to denote the output of component $X$ at layer $L$. %
The first layer takes as input a sequence of feature vectors $\langle \ExpectedArrivalTime, \DepotIncidentRate\rangle_{\SingleDepot}$ corresponding to each responder $v \in \RegionResponders$. 
After the first layer, the input of each subsequent layer is the output of the preceding layer. Accordingly, we can express $\textit{Input}_{\textnormal{MHA}}^L$ formally as~follows:
\begin{align*}
&\textit{Input}_{\textnormal{MHA}}^L = \\
    &~~~~~~~~~\begin{cases}
    \lbrace \langle \ExpectedArrivalTime, \DepotIncidentRate \;|\; \forall{\SingleDepot} \rangle \;|\;\forall{\SingleResponder} \rbrace& \text{if } \textit{L = 1} \\[5pt]
     \textit{Output}^{L-1}_{\textnormal{Norm-MLP}}           & \text{otherwise.}
\end{cases}
\end{align*}

We feed these values as inputs to MHA (i.e., \textit{query}, \textit{key}, \textit{value}): 
\begin{equation}
\textit{Output}^L_{\textnormal{MHA}} = \textbf{MHA}(\textit{Input}^L_{\textnormal{MHA}})
\label{eqn:trans_step_MHA}
\end{equation}
where MHA is a multi-head attention layer, same as \cite{vaswani2017attention}, where we set the dimension of key $|\textit{key}| = |\RegionDepots|$ (i.e., number of depots in the region $\SingleRegion$).

Then, we add the output $\textit{Output}^L_{\textnormal{MHA}}$ of the MHA with its input $\textit{Input}^L_{\textnormal{MHA}}$ and apply the normalization layer from \cite{ba2016layer}:
\begin{equation}
 \textit{Output}^L_{\textnormal{Norm-MHA}} = \textnormal{Norm}
 (\textnormal{Add}(\textit{Input}^L_{\textnormal{MHA}}, \textit{Output}^L_{\textnormal{MHA}}))
\label{eqn:trans_step_MHA_norm}
\end{equation}

Next, we feed the output of the normalization into an multilayer perceptron (MLP):
\begin{equation}
\textit{Output}^L_{\textnormal{MLP}} = \textbf{MLP}(    \textit{Output}^L_{\textnormal{Norm-MHA}})
\label{eqn:trans_step_MLP}
\end{equation}
Then, the output $\textit{Output}^L_{\textnormal{MLP}}$ of the MLP is added with the output $\textit{Output}^L_{\textnormal{Norm-MHA}}$ of the normalization after MHA, followed by another  normalization layer \cite{ba2016layer}:
\begin{equation}
 \textit{Output}^L_{\textnormal{Norm-MLP}} = \textnormal{Norm}(\textnormal{Add}
 (\textit{Output}^L_{\textnormal{Norm-MHA}}, \textit{Output}^L_{\textnormal{MLP}}))
\label{eqn:trans_step_MLP_norm}
\end{equation}

The steps indicated by \cref{eqn:trans_step_MHA,eqn:trans_step_MHA_norm,eqn:trans_step_MLP,eqn:trans_step_MLP_norm} are repeated $N$ times sequentially, and then we feed the output of the $N^{\textit{th}}$ layer into a softmax layer. 
We apply softmax separately to each responder $v \in \RegionResponders$ (i.e., we apply it separately to each element of the sequence of vectors, consisting of one vector for each responder $v \in \RegionResponders$) to obtain the likelihood of assigning responder $v$ to each depot in the region. Accordingly, we obtain $\LLPAction[v]$ as the output after applying softmax for each responder $v$:
\[
    \LLPAction[v] = \textit{Softmax}(  \textit{Output}^N_{\textnormal{Norm-MLP}}[v])
\]
ensuring that $\LLPAction[v] \cdot 1 = 1$ and $\LLPAction[v] \geq 0$ for each $v \in \RegionResponders$.

\subsection{Discretizing the Continuous Action}
\label{app:discrete_low_level}

We use maximum weight matching (MWM) in a weighted bipartite graph to efficiently compute a discrete assignment of responders to depots $\Assignments[\OneRegion]$ that is similar to the continuous actor output $\LLPAction$. There is a set of graph nodes representing responders $\RegionResponders$ and a set of nodes representing depots $\RegionDepots$; each responder $\SingleResponder$ is connected to each depot $\SingleDepot$ by an edge of weight $\LLPAction[v][d]$ (the low-level agent's preference for assigning $v$ to $d$). Maximizing with respect to the weights provides a discrete assignment close to the continuous action. 

We formally define the maximum weight matching to discretize a continuous action as follows.

First, we define the binary decision variable $x_{v,d} \in \{0, 1\}$ for $\forall{\SingleResponder}$ and $\forall{\SingleDepot}$, where
\[
 x_{v,d} = 
    \begin{cases}
    1& \text{if responder $v$ is assigned to depot $d$} \\[5pt]
     0& \text{otherwise.}
\end{cases}
\]

Next, a \textit{matching} in the graph assigns each responder to at most one depot. a \textit{maximum matching} assigns each responder to exactly one depot. We can \textit{matching} using following two constraints:

First, each responder ${\SingleResponder}$ in the region  needs to be assigned to one depot within the region:
\[
\sum_d^{\RegionDepots} x_{v,d} = 1
\]
Second, at most one responder can be assigned to each depot $\SingleDepot$ in the region:
\[
\sum_v^{\RegionResponders} x_{v,d} \leq 1
\]

Finally, we formulate the discrete assignment problem with the objective of finding feasible $\langle x_{v,d} \rangle$ that assign vehicles to depots with high $\LLPAction[v][d]$ (i.e., assign each vehicle $v$ to a depot $d$ for which the actor output a high likelihood $\LLPAction[v][d]$ of assignment).
Formally, the discrete assignment problem maximizes the following objective: 
\[
 \textnormal{Objective}_{\textnormal{MWM}} = \sum_d^{\RegionDepots} \sum_v^{\RegionResponders} x_{v,d} \cdot \LLPAction[v][d]
\]
where $\LLPAction[v][d]$ is the likelihood output by the actor for assigning responder $\SingleResponder$ to depot $\SingleDepot$.

The above problem is a \emph{maximum weight matching problem}, which is computationally trivial to solve using standard approaches (e.g., Edmonds' algorithm).

\section{High-Level Decision Agent}

\subsection{Discretizing the Continuous Action}
\label{app:redistribution}

\begin{algorithm}
\caption{$\textbf{GreedyAlgorithm}$}
 \label{algo:greedy_algo}

\textbf{Input}: $\NormalizedHLPAction, \Regions, \Depots, \Responders$\\
\textbf{Output}: $\TransHLPAction$

\begin{algorithmic}[1]
 
\STATE $V_{\textit{avail}} \leftarrow  |\Responders|$
 
\FOR {$\SingleRegion$}
{
\STATE $\TransHLPAction[\textit{g}] \leftarrow  0$
}
\ENDFOR

\WHILE{$\SumTransHLPAction[\textit{g}] < V_{\textit{avail}}$}
{

\FOR{$\SingleRegion$}
{
\STATE $\TransHLPAction[\textit{g}] \leftarrow \left\lfloor\frac{\NormalizedHLPAction[\textit{g}]}{\SumHatNormalizedHLPAction[\hat{\textit{g}}]} \cdot V_{\textit{avail}}\right\rfloor$

}
\ENDFOR

\STATE $V_{\textit{remain}} \leftarrow V_{\textit{avail}}  - \SumTransHLPAction[\textit{g}]$

\WHILE{$V_{\textit{remain}} > 0$}{

\STATE $\hat{\textit{g}} \leftarrow \argmax_{\SingleRegion} (\NormalizedHLPAction[\textit{g}] \cdot V_{\textit{avail}}   - \TransHLPAction[\textit{g}]) $

\STATE $\TransHLPAction[\hat{\textit{g}}] \leftarrow \TransHLPAction[\hat{\textit{g}}] + 1$

\STATE $V_{\textit{remain}} \leftarrow V_{\textit{remain}} - 1$
}
\ENDWHILE

\FOR{$\SingleRegion$}
{

\IF{ $\TransHLPAction[\textit{g}] >  |\RegionDepots|$}
{

\STATE $\TransHLPAction[\textit{g}] \leftarrow  |\RegionDepots|$

\STATE $\Regions \leftarrow \Regions \setminus \{\textit{g}\}$

\STATE $V_{\textit{avail}} \leftarrow V_{\textit{avail}} - |\RegionDepots|$
}
\ENDIF

}
\ENDFOR

}
\ENDWHILE

\end{algorithmic}
\end{algorithm}

In \cref{sec:high_level}, we provide a brief description of generating the discrete number of responders allocated to each region $\TransHLPAction$, where $\TransHLPAction = \argmin |\TransHLPAction - \NormalizedHLPAction|$, from a normalized continuous action $\NormalizedHLPAction$. Here, we explain the greedy algorithm (see \cref{algo:greedy_algo}) that we use to compute a feasible discrete allocation $\TransHLPAction$ that is similar to the continuous allocation $\NormalizedHLPAction$.
Please note that this algorithm is trivial (finding a vector of integer values subject to upper bounds, minimizing the difference to a  vector of desired continuous values); we provide a description for the sake of completeness.

We initialize the greedy algorithm with no vehicles allocated to any region. Then, we follow an iterative process, which includes three steps. First, we determine the allocation for every region based on the available responders $V_{\textit{avail}}$ and the ratio between the action value $\HLPAction[\textit{g}]$ corresponding to the region and the sum of all the action values $\SumHatHLPAction[\hat{\textit{g}}]$. Then, we compute the set of remaining responders ($V_{\textit{remain}}$) as the difference between all available responders $V_{\textit{avail}}$ and the responders allocated in previous steps $\SumTransHLPAction[\textit{g}]$. If there are any responders left awaiting allocation, we follow another iterative process. In each iteration, we choose the region with the highest difference between expected and allocated responders using the previous step (i.e., $\NormalizedHLPAction[\textit{g}] \cdot V_{\textit{avail}}   - \TransHLPAction[\textit{g}]$). Then, we add one more responder to the chosen region. Finally, we check if the allocation to any region exceeds the number of depots in the region; in case of such a situation, the allocation is fixed at the number of depots (to avoid allocating more responders than what is feasible), and the region is removed from future iterations.

\subsection{Assignment of Responders to Depots}
\label{app:high_level_assignment}

Next, we discuss how the high-level agent decides which responders should leave each region (for regions whose allocations have been reduced) and to which depots these responders should be assigned in their new regions (which are regions whose allocations have been increased).
To minimize the gap in coverage while responders drive to their new regions, we formulate this assignment as a minimum cost flow problem (MCFP) that minimizes the total travel time of all the reallocated responders.

\paragraph{Formulation of Minimum Cost Flow Problem}
To formulate the assignment problem as an MCFP, we define a graph $\textbf{G}(\Nodes, \Edges)$ with nodes $\Nodes$ and edges $\Edges$. Each node $\SingleNode$ represents one of the following: 
\begin{itemize}
    \item [1.)] abstract source $\SSNodes_{\textit{\tiny{source}}}$ and sink $\SSNodes_{\textit{\tiny{sink}}}$ nodes for the flow
    \item [2.)] regions $\Regions_{\textit{\tiny{leaves}}}$ from which responders will leave: ${\Regions_{\textit{\tiny{leaves}}}} = \{\OneRegion \mid \SingleRegion, \TransPrevHLPAction[\OneRegion] > \TransHLPAction[\OneRegion] \}$
    \item [3.)] regions $\Regions_{\textit{arrive}}$ where responders will arrive: ${\Regions_{\textit{arrive}}} = \{\OneRegion \mid \SingleRegion, \TransPrevHLPAction[\OneRegion] < \TransHLPAction[\OneRegion] \}$
    \item [4.)] all the responders $\Responders_{\textit{\tiny{leaves}}}$ in  regions $\OneRegion \in {\Regions_{\textit{\tiny{leaves}}}}$ from which responders will leave: $\Responders_{\textit{\tiny{leaves}}} = \bigcup_\OneRegion^{\Regions_{\textit{\tiny{leaves}}}} \RegionResponders$
    \item [5.)] unoccupied depots $\Depots_{\textit{\tiny{unoccupied}}}$ in regions $\OneRegion \in \Regions_{\textit{arrive}}$ where responders will arrive: $\Depots_{\textit{\tiny{unoccupied}}} = \bigcup_\OneRegion^{\Regions_{\textit{arrive}}} \lbrace d \mid d \in \RegionDepots \land d \notin  \{d^v \mid v \in \RegionResponders \} \rbrace$
\end{itemize}

A pair of two nodes $y_1, y_2 \in \Nodes$ are connected by a directed edge $(y_1, y_2) \in \Edges$ if and only if one of the following conditions met:
\begin{itemize}
    \item [1.)] $y_1 = \SSNodes_{\textit{\tiny{source}}} \land y_2 \in {\Regions_{\textit{\tiny{leaves}}}}$: the source node is connected to all nodes representing the regions from which responders will leave; 
    \item [2.)] $y_1 = \OneRegion^{y_2} \land y_2 \in \Responders_{\textit{\tiny{leaves}}}$:
    each node representing a region from which responders will leave is connected to all nodes that represent the responders in that region; 
    \item [3.)] $y_1 \in \Responders_{\textit{\tiny{leaves}}} \land y_2 \in {\Depots_{\textit{\tiny{unoccupied}}}}$: each node representing a responder from a region from which responders will leave is connected to all nodes representing unoccupied depots in regions where responders will arrive; 
    \item [4.)] $y_1 \in {\Depots_{\textit{\tiny{unoccupied}}}} \land y_2 \in {\Regions_{\textit{\tiny{leaves}}}} \land y_1 \in \Depots^{y_2}$: each node that represents an unoccupied depot in a region where responders will arrive is connected to the node representing the region of the depot; 
    \item [5.)] $y_1 \in {\Regions_{\textit{\tiny{leaves}}}} \land  y_2 = \SSNodes_{\textit{\tiny{sink}}}$: all nodes representing regions where responders will arrive are connected to the sink node (see \cref{fig:high_level_process} for an illustration).
\end{itemize}

Each edge $(y_1, y_2) \in \Edges$ has a cost $\textbf{Cost}(y_1,y_2)$ and a capacity $c(y_1,y_2)$. We let the cost between nodes representing $v \in {\Responders_{\textit{\tiny{leaves}}}}$ and $d \in {\Depots_{\textit{\tiny{unoccupied}}}}$ be the total time $\ExpectedArrivalTime$ that it would take  responder $v$ to move to  depot $d$; and for all other edges, we let the cost be zero. We let the capacity $c(y_1,y_2)$ of directed edges $(y_1, y_2) \in \Edges$ be the following:
\[
c(y_1,y_2) = 
\begin{cases}
\TransPrevHLPAction[\OneRegion] - \TransHLPAction[\OneRegion] & \text{if }   y_1 \in \SSNodes_{\textit{\tiny{source}}} \land y_2 \in {\Regions_{\textit{\tiny{leaves}}}}\\
\TransHLPAction[\OneRegion] - \TransPrevHLPAction[\OneRegion]  & \text{if }   y_1 \in {\Regions_{\textit{\tiny{arrives}}}} \land y_2 \in \SSNodes_{\textit{\tiny{sink}}}\\
1              & \text{otherwise.}
\end{cases}
\]

Finally, we require the total amount of flow from source $\SSNodes_{\textit{\tiny{source}}}$ to sink  $\SSNodes_{\textit{\tiny{sink}}}$ to be $\sum_\OneRegion^{{\Regions_{\textit{\tiny{leaves}}}}} (\TransPrevHLPAction[\OneRegion] - \TransHLPAction[\OneRegion])$.

\paragraph{Minimum Cost Flow as a Responder Assignment}
The integer solution of the above minimum cost flow problem is an assignment that minimizes the total travel time of the reallocated responders.
First, for each responder $v$, an integer solution of the above problem assigns a positive flow for at most one depot $d$.
This assignment is feasible:
for each region from which responders will leave, the right number of responders will have a positive flow; and for each region where responders will arrive, the right number of depots will have a positive flow.
If each responder $v$ drives to the assigned depot $d$, then their total travel time will be minimal.
Finding an integer solution to a minimum cost flow problem is computationally easy.

\paragraph{Standard Capacity and Conservation Constraints}
The above formulation is subject to the standard constraints of flow conservation and edge capacity, and its objective is standard minimization of flow cost.
For the sake of completeness, we provide a formal specification of these constraints and the objective.
Let integer variable $x(y_1,y_2)$ indicate whether there is flow from node $y_1 \in \Nodes$ to node $y_2 \in \Nodes$ (i.e., if $x(y_1,y_2) = 0$ then there is no flow, meaning possible responder movement, otherwise there is flow). In addition, the formulation needs to ensure the following flow constraints. First, the net flow in all nodes, except the source and sink nodes, must be equal to zero:
\[
    \sum_{j \in \Nodes} x(j,i) - \sum_{k \in \Nodes} x(i,k)= 0 \;\;\;\; \forall{i \in \Nodes \setminus \{\SSNodes_{\textit{\tiny{source}}}, \SSNodes_{\textit{\tiny{sink}}}\}}
\]

Second, responders that leave their current region must arrive at a new region: 
\[
      \sum_{i \in \Nodes} x(\SSNodes_{\textit{\tiny{source}}},i) = \sum_{j \in \Nodes} x(j,\SSNodes_{\textit{\tiny{sink}}}) = \sum_\OneRegion^{{\Regions_{\textit{\tiny{leaves}}}}} (\TransPrevHLPAction[\OneRegion] - \TransHLPAction[\OneRegion])
\]

Third, each edge $(y_1,y_2) \in \Edges$ can only accommodate responders up to the capacity of the edge $ x(y_1,y_2) \leq c(y_1,y_2)$. Since the cost is zero for edges that do not connect a responder at one end with a depot at the other end (i.e., \textbf{Cost}$(y_1,y_2) = 0$, if $y_1 \notin  {\Responders_{\textit{\tiny{leaves}}}} \lor y_2 \notin {\Depots_{\textit{\tiny{unoccupied}}}}$), we express the objective of the formulation based on the following expression:
\[
    \textnormal{Objective}_{\textnormal{MCFP}} =  \min  \sum_v^{\Responders_{\textit{\tiny{leaves}}}} \sum_d^{\Depots_{\textit{\tiny{unoccupied}}}} \ExpectedArrivalTime \cdot x(v, d)
\]

\section{Additional Numerical Results}

\subsection{Description of Baselines}
\label{app:baseline_description}

Due to limited space, we described some baselines briefly in the main text. Here, we provide detailed descriptions. %

\paragraph{$p$-Median-Based Policy} 
For each region, the low-level responder reallocation problem is mapped to a $p$\emph{-median problem}, which assigns the responders to depots in the region %
so that the average demand-weighted distance between cells and the nearest depots is minimized. 
However, the $p$-median formulation does not account for responders being unavailable while serving incidents. \cite{vazirizade2021learning} modify the standard formulation by including a balancing term $\alpha$ in the objective, which penalizes responders that cover areas with disproportionately more incidents compared to other responders. If $\alpha = 0$, then the problem is the standard $p$-median problem; and $\alpha > 0$ penalizes responders that cover areas with disproportionally higher incident~rates. 

\paragraph{Greedy Policy} The greedy policy is a simple heuristic approach that reallocates responders in a region to depots based the incident rates and the expected travel times to the depots, similar to the features that we use in our learning-based approach. %
Comparison to this heuristic approach as a baseline demonstrates the need for a more complex, learning-based approach instead of simple heuristics based on basic features, such as incident rates around depots and expected travel times. 
At each decision epoch,
the greedy algorithm first chooses the $|\RegionResponders|$ depots in  region~$\OneRegion$ that have the highest nearby incident rates ($\DepotIncidentRate$).
Then, the greedy algorithm reallocates the responders to these depots using minimum-weight perfect matching based on the expected travel time between depot locations and the current positions of the responders as weights (i.e., minimizing total travel times for reallocation).

\subsection{Additional Numerical Results for Nashville}
\label{app:additional_nashville}

For the DRLSN baseline~\cite{ji2019deep}, we adapt the centralized approach to a hierarchical one similar to \cite{pettet2021hierarchical2} and use the same high-level planner as \cite{pettet2021hierarchical2} (same approach that we follow for the other baselines, such as $p$-median based policy, greedy policy, and static policy). The rationale behind this extension is to provide all approaches with the benefit of hierarchical planning; otherwise, DRLSN would inherently be at a disadvantage. For our experiments with DRLSN, we consider triggering both the high-level planner and the low-level planners every time an incident arrives or when there have been no incidents in the past 60 minutes, same as in \cite{pettet2021hierarchical2}. 

\begin{figure*}
\centering
\begin{subfigure}{0.3\textwidth}
\pgfplotstableread[col sep=comma,]{data_nashville/city_level_summary_5_24.csv}\regionfive
\pgfplotstableread[col sep=comma,]{data_nashville/city_level_summary_6_24.csv}\regionsix
\pgfplotstableread[col sep=comma,]{data_nashville/city_level_summary_7_24.csv}\regionseven
\begin{tikzpicture}
\begin{axis}[
      boxplot/draw direction=y,
      width=\columnwidth,
      xtick={1,2,3},
      xticklabels={{5}, {6}, {7}},
      height = 4.0cm,
      ymajorgrids,
      major grid style={draw=gray!25},
      bugsResolvedStyle/.style={},
      ylabel={Average Response Time [s]},
      xlabel={Number of Regions},
      font=\small,
    ]

\addplot+[boxplot={box extend=0.08, draw position=1}, ColorVPG, solid, rshift, fill=ColorVPG!20, mark=x] table [col sep=comma, y=vpg] {\regionfive};
\addplot+[boxplot={box extend=0.08, draw position=1}, ColorMCTS, solid, fill=ColorMCTS!20, mark=x] table [col sep=comma, y=mcts] {\regionfive};
\addplot+[boxplot={box extend=0.08, draw position=1}, ColorOur, solid, lshift, fill=ColorOur!20, mark=x] table [col sep=comma, y=drl] {\regionfive};

\addplot+[boxplot={box extend=0.08, draw position=2}, ColorVPG, solid, rshift, fill=ColorVPG!20, mark=x] table [col sep=comma, y=vpg] {\regionsix};
\addplot+[boxplot={box extend=0.08, draw position=2}, ColorMCTS, solid, fill=ColorMCTS!20, mark=x] table [col sep=comma, y=mcts] {\regionsix};
\addplot+[boxplot={box extend=0.08, draw position=2}, ColorOur, solid, lshift, fill=ColorOur!20, mark=x] table [col sep=comma, y=drl] {\regionsix};

\addplot+[boxplot={box extend=0.08, draw position=3}, ColorVPG, solid, rshift, fill=ColorVPG!20, mark=x] table [col sep=comma, y=vpg] {\regionseven};
\addplot+[boxplot={box extend=0.08, draw position=3}, ColorMCTS, solid, fill=ColorMCTS!20, mark=x] table [col sep=comma, y=mcts] {\regionseven};
\addplot+[boxplot={box extend=0.08, draw position=3}, ColorOur, solid, lshift, fill=ColorOur!20, mark=x] table [col sep=comma, y=drl] {\regionseven};

  \end{axis}
\end{tikzpicture}
\caption{24 Responders
}
\end{subfigure}\hfill
\begin{subfigure}{0.3\textwidth}
\pgfplotstableread[col sep=comma,]{data_nashville/city_level_summary_5_26.csv}\regionfive
\pgfplotstableread[col sep=comma,]{data_nashville/city_level_summary_6_26.csv}\regionsix
\pgfplotstableread[col sep=comma,]{data_nashville/city_level_summary_7_26.csv}\regionseven

\begin{tikzpicture}
\begin{axis}[
      boxplot/draw direction=y,
            width=\columnwidth,
      xtick={1,2,3},
      xticklabels={{5}, {6}, {7}},
      height = 4.0cm,
      ymajorgrids,
      major grid style={draw=gray!25},
      bugsResolvedStyle/.style={},
      ylabel={Average Response Time [s]},
      xlabel={Number of Regions},
      font=\small,
    ]

\addplot+[boxplot={box extend=0.08, draw position=1}, ColorVPG, solid, rshift, fill=ColorVPG!20, mark=x] table [col sep=comma, y=vpg] {\regionfive};
\addplot+[boxplot={box extend=0.08, draw position=1}, ColorMCTS, solid, fill=ColorMCTS!20, mark=x] table [col sep=comma, y=mcts] {\regionfive};
\addplot+[boxplot={box extend=0.08, draw position=1}, ColorOur, solid, lshift, fill=ColorOur!20, mark=x] table [col sep=comma, y=drl] {\regionfive};

\addplot+[boxplot={box extend=0.08, draw position=2}, ColorVPG, solid, rshift, fill=ColorVPG!20, mark=x] table [col sep=comma, y=vpg] {\regionsix};
\addplot+[boxplot={box extend=0.08, draw position=2}, ColorMCTS, solid, fill=ColorMCTS!20, mark=x] table [col sep=comma, y=mcts] {\regionsix};
\addplot+[boxplot={box extend=0.08, draw position=2}, ColorOur, solid, lshift, fill=ColorOur!20, mark=x] table [col sep=comma, y=drl] {\regionsix};

\addplot+[boxplot={box extend=0.08, draw position=3}, ColorVPG, solid, rshift, fill=ColorVPG!20, mark=x] table [col sep=comma, y=vpg] {\regionseven};
\addplot+[boxplot={box extend=0.08, draw position=3}, ColorMCTS, solid, fill=ColorMCTS!20, mark=x] table [col sep=comma, y=mcts] {\regionseven};
\addplot+[boxplot={box extend=0.08, draw position=3}, ColorOur, solid, lshift, fill=ColorOur!20, mark=x] table [col sep=comma, y=drl] {\regionseven};

\end{axis}
\end{tikzpicture}
\caption{26 Responders}
\end{subfigure}\hfill
\begin{subfigure}{0.3\textwidth}
\pgfplotstableread[col sep=comma,]{data_nashville/city_level_summary_5_28.csv}\regionfive
\pgfplotstableread[col sep=comma,]{data_nashville/city_level_summary_6_28.csv}\regionsix
\pgfplotstableread[col sep=comma,]{data_nashville/city_level_summary_7_28.csv}\regionseven

\begin{tikzpicture}
\begin{axis}[
      boxplot/draw direction=y,
            width=\columnwidth,
      xtick={1,2,3},
      xticklabels={{5}, {6}, {7}},
      height = 4.0cm,
      ymajorgrids,
      major grid style={draw=gray!25},
      bugsResolvedStyle/.style={},
      ylabel={Average Response Time [s]},
      xlabel={Number of Regions},
      font=\small,
    ]

\addplot+[boxplot={box extend=0.08, draw position=1}, ColorVPG, solid, rshift, fill=ColorVPG!20, mark=x] table [col sep=comma, y=vpg] {\regionfive};
\addplot+[boxplot={box extend=0.08, draw position=1}, ColorMCTS, solid, fill=ColorMCTS!20, mark=x] table [col sep=comma, y=mcts] {\regionfive};
\addplot+[boxplot={box extend=0.08, draw position=1}, ColorOur, solid, lshift, fill=ColorOur!20, mark=x] table [col sep=comma, y=drl] {\regionfive};

\addplot+[boxplot={box extend=0.08, draw position=2}, ColorVPG, solid, rshift, fill=ColorVPG!20, mark=x] table [col sep=comma, y=vpg] {\regionsix};
\addplot+[boxplot={box extend=0.08, draw position=2}, ColorMCTS, solid, fill=ColorMCTS!20, mark=x] table [col sep=comma, y=mcts] {\regionsix};
\addplot+[boxplot={box extend=0.08, draw position=2}, ColorOur, solid, lshift, fill=ColorOur!20, mark=x] table [col sep=comma, y=drl] {\regionsix};

\addplot+[boxplot={box extend=0.08, draw position=3}, ColorVPG, solid, rshift, fill=ColorVPG!20, mark=x] table [col sep=comma, y=vpg] {\regionseven};
\addplot+[boxplot={box extend=0.08, draw position=3}, ColorMCTS, solid, fill=ColorMCTS!20, mark=x] table [col sep=comma, y=mcts] {\regionseven};
\addplot+[boxplot={box extend=0.08, draw position=3}, ColorOur, solid, lshift, fill=ColorOur!20, mark=x] table [col sep=comma, y=drl] {\regionseven};

\end{axis}
\end{tikzpicture}
\caption{28 Responders}
\end{subfigure}

\caption{Distribution of average response times (lower is better) with our approach (\textcolor{ColorLegendOur}{$\blacksquare$}), MCTS (\textcolor{ColorLegendMCTS}{$\blacksquare$}), and DRLSN (\textcolor{ColorLegendVPG}{$\blacksquare$}) for 10 different sample incident chains (\textbf{Nashville}).
In this figure, we plot the same data for our approach and MCTS as in \cref{fig:city_level_response_times_twenties_nashville_twenty_four,fig:city_level_response_times_twenties_nashville_twenty_six,fig:city_level_response_times_twenties_nashville_twenty_eight}; the only difference is the inclusion of DRLSN, which changes the scaling of the vertical axis.
}
\label{fig:city_level_response_times_twenties_full_nashville}
\end{figure*}
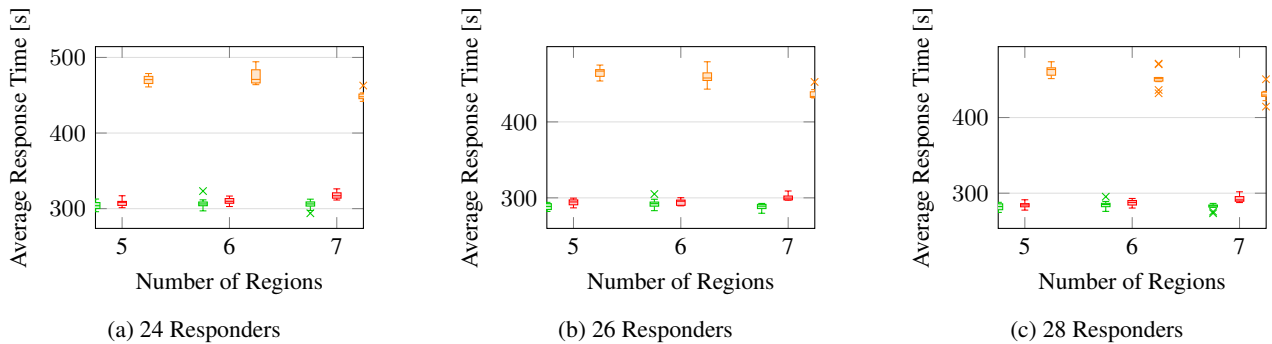

\cref{fig:city_level_response_times_twenties_full_nashville} shows the average response times for our approach, MCTS, and DRLSN based on the same sample chains that we used in the response-time analysis in the numerical results of the main text (see \cref{subsec:evaluation}). We observe that  in all cases, DRLSN suffers from response times higher than around 450, which is 50\% worse than our approach. Note that we use the same data here as in Figures 4a to 4c (in the main text); the  difference in the scaling of the vertical axis between the two sets of figures is due to the inclusion of DRSLN as a baseline (which performs significantly worse than MCTS and the proposed DDPG-based approach, thereby changing the scale of the vertical axis).

\paragraph{Learning Curves}

\pgfplotstableread[col sep=comma,]{data_nashville/nashville_7_regions_city.csv}\regionsevenregioncity
\pgfplotstableread[col sep=comma,]{data_nashville/nashville_7_regions_region_0.csv}\regionsevenregionzero
\pgfplotstableread[col sep=comma,]{data_nashville/nashville_7_regions_region_5.csv}\regionsevenregionfive
\pgfplotstableread[col sep=comma,]{data_nashville/nashville_7_regions_region_6.csv}\regionsevenregionsix

\begin{figure*}[th]
\begin{subfigure}{0.32\textwidth}
\begin{tikzpicture}
\begin{axis}[
      width=\linewidth,
      height = 5cm,
      bugsResolvedStyle/.style={},
      ylabel={Response Time [Seconds]},
      xlabel={Training Episodes},
    ]

\addplot[very thick, ColorRandom] table [x=index, y=random_mean, col sep=comma] {\regionsevenregionzero};

\addplot[very thick, ColorRandom!50, name path=A] table [x=index, y=random_min, col sep=comma] {\regionsevenregionzero};
\addplot[very thick, ColorRandom!50, name path=B] table [x=index, y=random_max, col sep=comma] {\regionsevenregionzero};

\addplot[fill=ColorRandom, opacity=0.25] fill between[of=A and B];

\addplot[very thick, ColorOur] table [x=index, y=response_time_mean, col sep=comma] {\regionsevenregionzero};

\addplot[very thick, ColorOur!50, name path=A] table [x=index, y=response_time_min, col sep=comma] {\regionsevenregionzero};
\addplot[very thick, ColorOur!50, name path=B] table [x=index, y=response_time_max, col sep=comma] {\regionsevenregionzero};

\addplot[fill=ColorOur, opacity=0.25] fill between[of=A and B];
\end{axis}
\end{tikzpicture}
\caption{Region \#1}
\label{fig:nashville_region_1_learning_curve}
\end{subfigure}%
\begin{subfigure}{0.32\textwidth}
\begin{tikzpicture}
\begin{axis}[
      width=\linewidth,
      height = 5cm,
      bugsResolvedStyle/.style={},
      ylabel={Response Time [Seconds]},
      xlabel={Training Episodes},
    ]

\addplot[very thick, ColorRandom] table [x=index, y=random_mean, col sep=comma] {\regionsevenregionfive};

\addplot[very thick, ColorRandom!50, name path=A] table [x=index, y=random_min, col sep=comma] {\regionsevenregionfive};
\addplot[very thick, ColorRandom!50, name path=B] table [x=index, y=random_max, col sep=comma] {\regionsevenregionfive};

\addplot[fill=ColorRandom, opacity=0.25] fill between[of=A and B];

\addplot[very thick, ColorOur] table [x=index, y=response_time_mean, col sep=comma] {\regionsevenregionfive};

\addplot[very thick, ColorOur!50, name path=A] table [x=index, y=response_time_min, col sep=comma] {\regionsevenregionfive};
\addplot[very thick, ColorOur!50, name path=B] table [x=index, y=response_time_max, col sep=comma] {\regionsevenregionfive};

\addplot[fill=ColorOur, opacity=0.25] fill between[of=A and B];
\end{axis}
\end{tikzpicture}
\caption{Region \#6}
\label{fig:nashville_region_6_learning_curve}
\end{subfigure}%
\begin{subfigure}{0.32\textwidth}
\begin{tikzpicture}
\begin{axis}[
      width=\linewidth,
      height = 5cm,
      bugsResolvedStyle/.style={},
      ylabel={Response Time [Seconds]},
      xlabel={Training Episodes},
    ]

\addplot[very thick, ColorRandom] table [x=index, y=random_mean, col sep=comma] {\regionsevenregionsix};

\addplot[very thick, ColorRandom!50, name path=A] table [x=index, y=random_min, col sep=comma] {\regionsevenregionsix};
\addplot[very thick, ColorRandom!50, name path=B] table [x=index, y=random_max, col sep=comma] {\regionsevenregionsix};

\addplot[fill=ColorRandom, opacity=0.25] fill between[of=A and B];

\addplot[very thick, ColorOur] table [x=index, y=response_time_mean, col sep=comma] {\regionsevenregionsix};

\addplot[very thick, ColorOur!50, name path=A] table [x=index, y=response_time_min, col sep=comma] {\regionsevenregionsix};
\addplot[very thick, ColorOur!50, name path=B] table [x=index, y=response_time_max, col sep=comma] {\regionsevenregionsix};

\addplot[fill=ColorOur, opacity=0.25] fill between[of=A and B];
\end{axis}
\end{tikzpicture}
\caption{Region \#7}
\label{fig:nashville_region_7_learning_curve}
\end{subfigure}
\caption{Evolution of the performance of the low-level policy $\mu_{\OneRegion}$ throughout the training process for regions 1, 6, 7 of the 7-region decomposition in \textbf{Nashville}, measured as the average response time. The dark green line (\textcolor{ColorLegendOurThick}{$\blacksquare$})  indicates the average of 10 different policies (trained on the given number of episodes), which are evaluated on 5 different sample chains with the number of responders ranging from 1 to $|\RegionDepots|$. The light green area (\textcolor{ColorLegendOur}{$\blacksquare$}) indicates the 10th to 90th percentiles of average response times  over 10 different policies. The dark gray line (\textcolor{ColorLegendRandomThick}{$\blacksquare$})  indicates the mean of average response time over the same set of samples when using a random policy, and the light gray area (\textcolor{ColorLegendRandom}{$\blacksquare$}) indicates the 10th to 90th percentiles of average response times over 15 different random policies.}
\end{figure*}

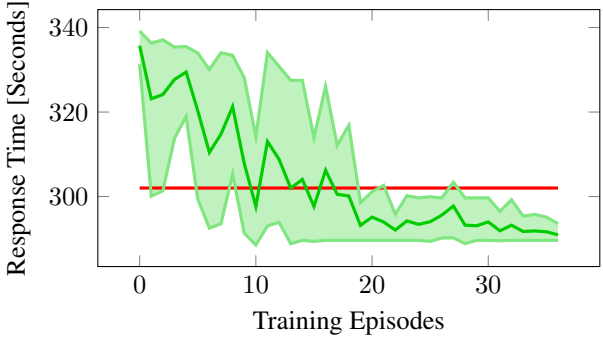
\begin{figure}[t]
\begin{tikzpicture}
\begin{axis}[
      width=\linewidth,
      height = 5cm,
      bugsResolvedStyle/.style={},
      ylabel={Response Time [Seconds]},
      xlabel={Training Episodes},
    ]

\addplot[very thick, ColorMCTS] table [x=index, y=mcts, col sep=comma] {\regionsevenregioncity};

\addplot[very thick, ColorOur] table [x=index, y=response_time_mean, col sep=comma] {\regionsevenregioncity};

\addplot[very thick, ColorOur!50, name path=A] table [x=index, y=response_time_min, col sep=comma] {\regionsevenregioncity};
\addplot[very thick, ColorOur!50, name path=B] table [x=index, y=response_time_max, col sep=comma] {\regionsevenregioncity};

\addplot[fill=ColorOur, opacity=0.25] fill between[of=A and B];
\end{axis}
\end{tikzpicture}
\caption{Evolution of the performance of the policy $\mu_{h}$ throughout the training process for the high-level decision agent of the 7-region decomposition in \textbf{Nashville}, measured as the average response time. The dark green line (\textcolor{ColorLegendOurThick}{$\blacksquare$}) indicates the average of 5 different policies (trained on the given number of episodes), which are evaluated on 5 different sample chains with the number of responders in the entire city ranging from 24 to 28. The light green area (\textcolor{ColorLegendOur}{$\blacksquare$}) indicates the 10th to 90th percentiles of average response times over 5 different policies. The red line (\textcolor{ColorLegendMCTSThick}{$\blacksquare$}) indicates the average response times obtained when using the state-of-the-art MCTS approach on the same 5 sample chains with the number of responders in the entire city ranging from 24 to 28.}
\label{fig:nashville_city_learning_curve}
\end{figure}

\cref{fig:nashville_region_1_learning_curve,fig:nashville_region_6_learning_curve,fig:nashville_region_7_learning_curve} show the evolution of the low-level policy $\mu_{g}$ for regions 1, 6 and 7 of the 7-region decomposition in Nashville. 
The vertical axis indicates the average response times, and the horizontal axis indicates the number of training episodes. The light green area (\textcolor{ColorLegendOur}{$\blacksquare$}) represents the  10th to 90th  percentiles of average response times when evaluated over 5 different sample chains using 10 policies.
The light gray area (\textcolor{ColorLegendRandom}{$\blacksquare$})  represents the 10th to 90th percentiles of average response times over 15 different random policies. We observe that the trained policy performs considerably better than a random policy, even after a relatively low number of episodes.

\cref{fig:nashville_city_learning_curve} shows the performance of the policy $\mu_{h}$ for the high-level decision agent of the 7-region decomposition in Nashville. The vertical axis indicates the average response times, and the horizontal axis indicates the number of training episodes. The light green area represents the 10th to 90th percentiles of the average response times obtained at each training episode for 5 different policies. Based on the results, our high-level policy training helps to reduce the response time compared to the state-of-the-art approach.

\paragraph{Noisy Observations}
We perform experiments to verify whether our policies are resilient to noisy observations during evaluation, which can model uncertainty in predicting incident rates and travel times. As the observation values are non-negative by definition, we employ a multiplicative noise drawn from a log-normal distribution with zero mean and standard deviation ranging from 0.1 to 0.3 (note these are the mean and std. dev. values for the normal distribution). These standard deviations roughly correspond to noise levels ranging from $\pm$7\% to $\pm$20\%.

We add noise to both low-level agent observations (Arrival time and Nearby incident rate) and high-level agent observations (Region incident rates, i.e., sum of incident rates for all the cells in each region). 

\begin{figure}
    \centering
\includegraphics[width=\columnwidth]{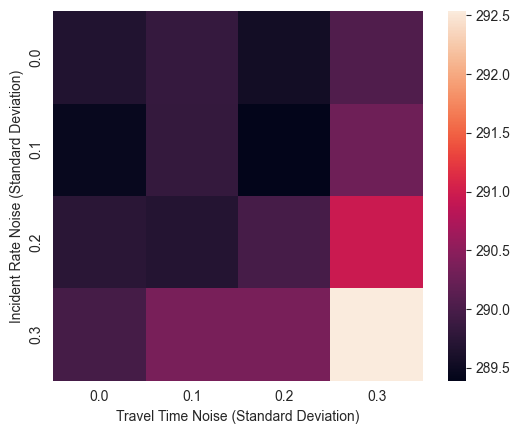}
    \caption{Caption}
    \label{fig:nashville_noise}
\end{figure}

In \cref{fig:nashville_noise}, each square shows the average response time for 10 sample chains (same as the ones used in the experimental section of our paper) for three different responder values (24, 26, and 28) with 7-region decomposition. The horizontal and vertical axes show the standard deviations of the long-normal noise for travel times and for incident rates, respectively. 

We observe that the increase in average response times is not significant even with $\pm$20\% noise added to both observed incident rates and travel times, demonstrating that our policies are robust to uncertain predictions of incident rates and travel times.

\subsection{Numerical Results for Seattle}
\label{app:seatle_numerical}

\begin{figure*}
\centering
\begin{subfigure}{0.3\textwidth}
\includegraphics[width=\textwidth]{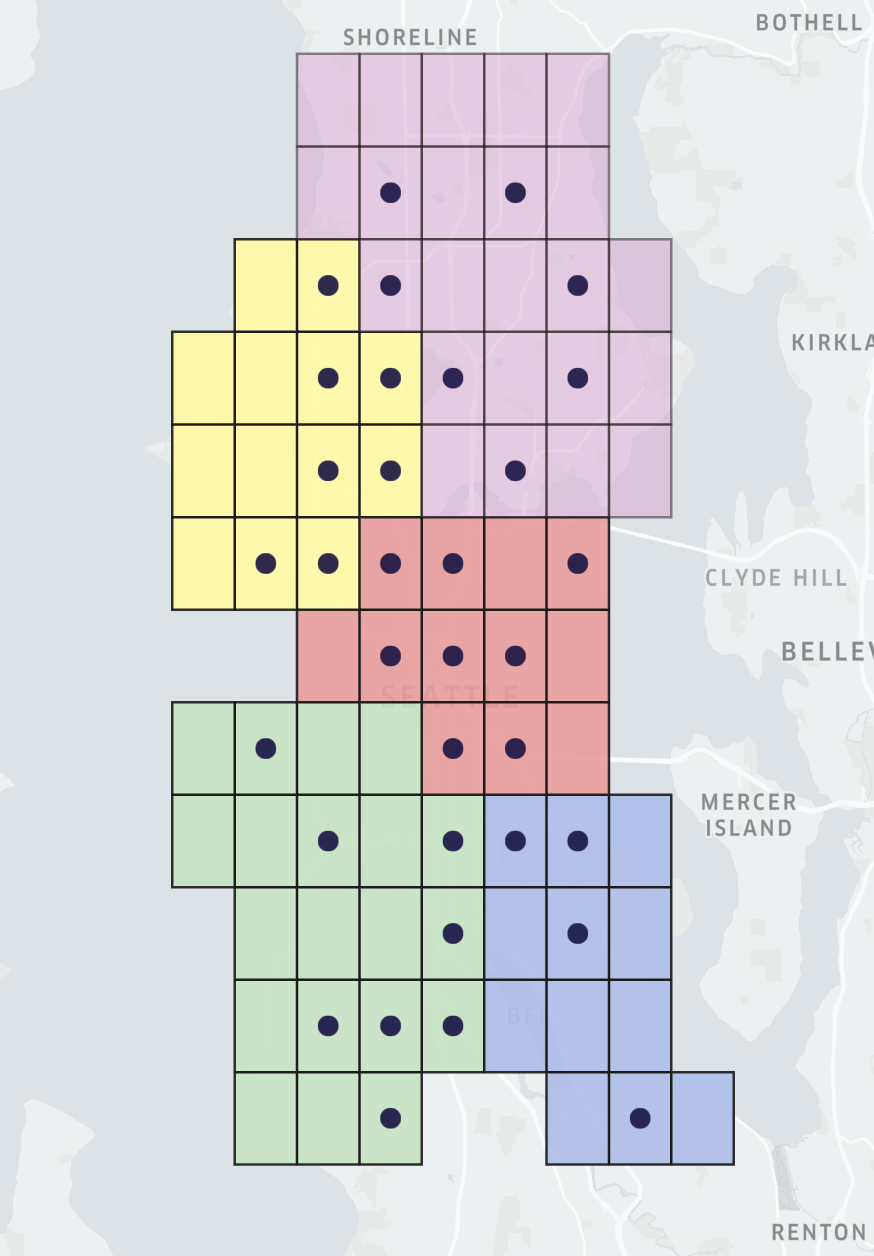}
\caption{5 Regions}
\end{subfigure}\hfill
\begin{subfigure}{0.3\textwidth}
\includegraphics[width=\textwidth]{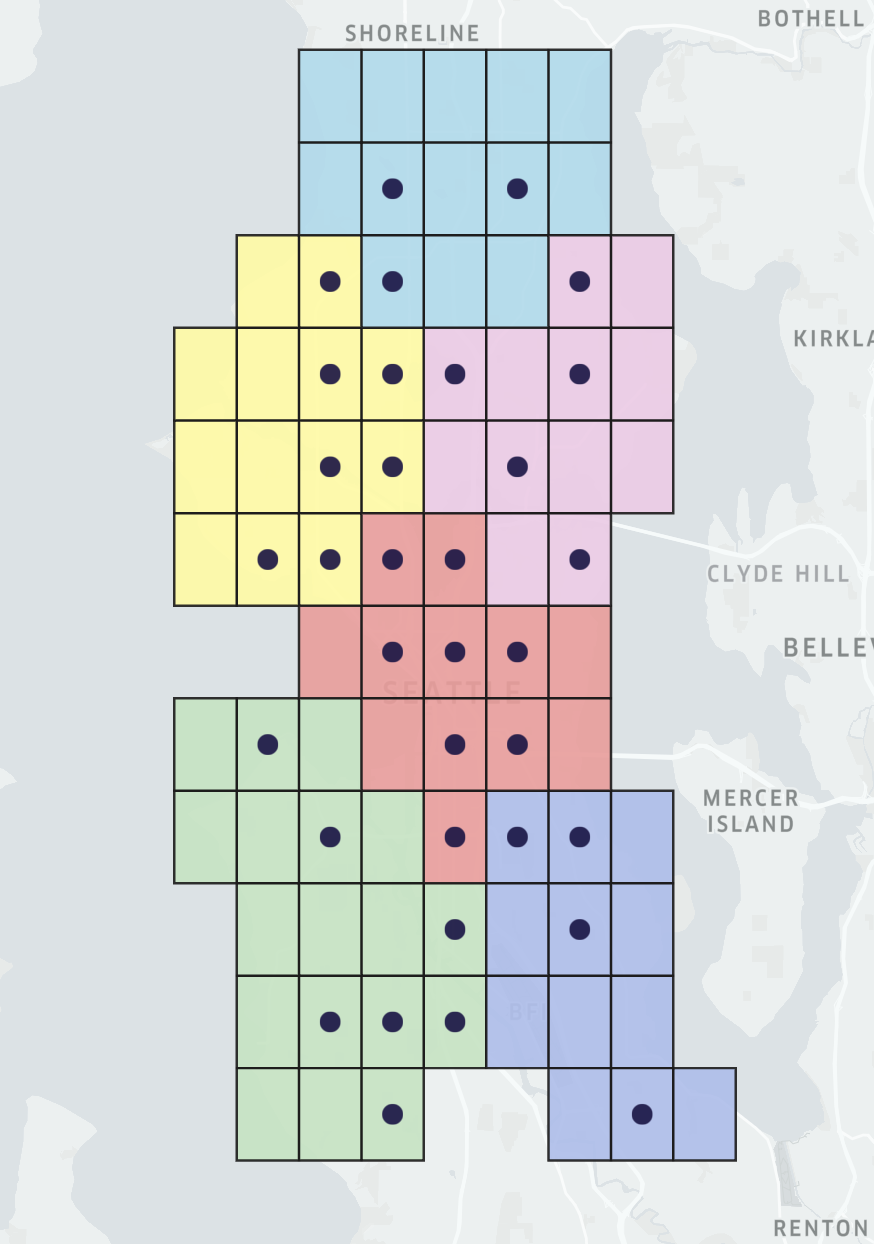}
\caption{6 Regions}
\end{subfigure}\hfill
\begin{subfigure}{0.3\textwidth}
\includegraphics[width=\textwidth]{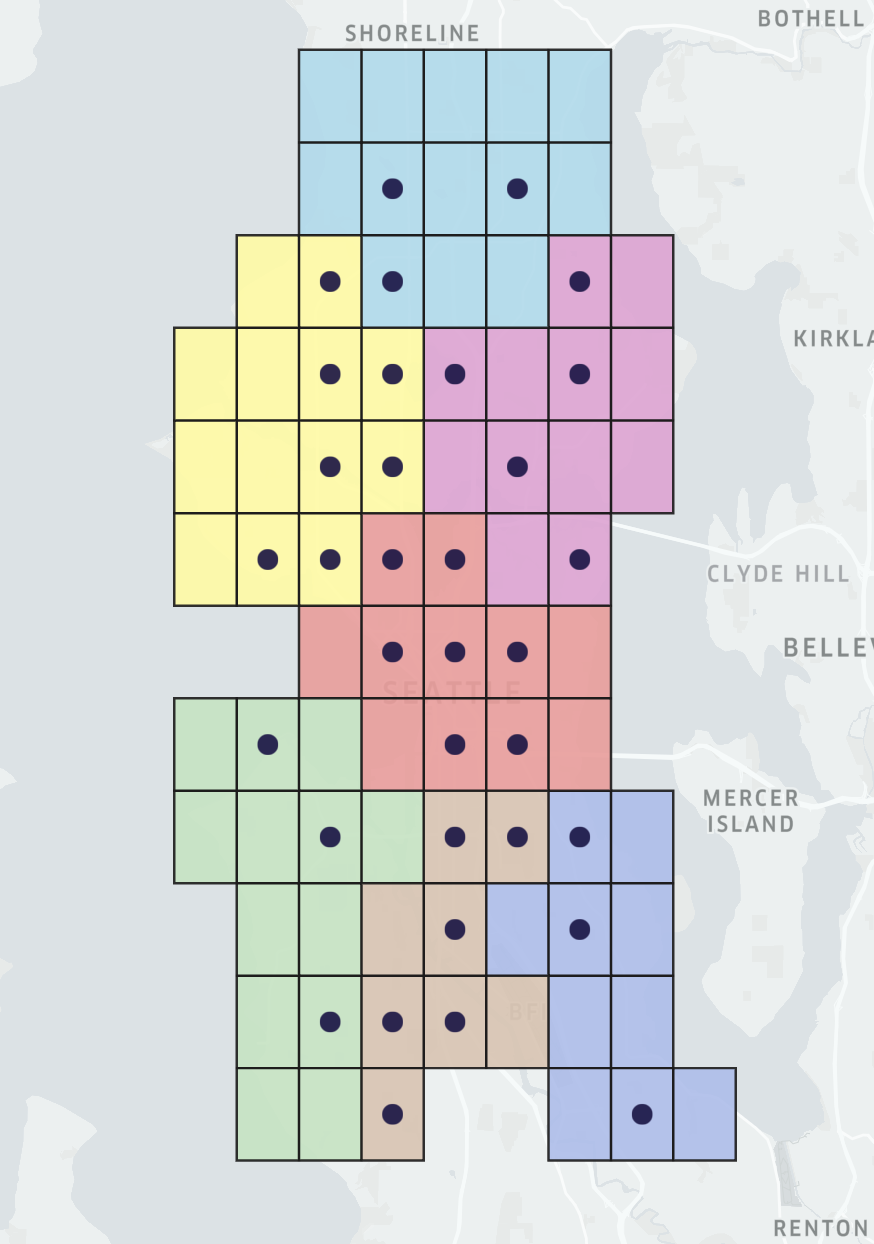}
\caption{7 Regions}
\end{subfigure}
\caption{Segmentation of Seattle into 5, 6, and 7 regions. Dots on the map indicate depots where responders can wait.}
\label{fig:clustering_seatle}
\end{figure*}

We also evaluate our algorithm using publicly available data from the U.S. city of Seattle, WA~\cite{seattleData}. 
We apply the same one-by-one mile square grid to the city to ensure the same experimental setup for the two geographical areas that we consider (i.e., for Nashville and Seattle). The city has 34 depots and 14 general hospitals to assist emergency response management. \cref{fig:clustering_seatle} shows the region segmentation for Seattle data (similar figures for Nashville data can be found in \cite{pettet2021hierarchical2}). We use  60 incident chains sampled from historical incident data distribution.  Each chain is sampled over 9 days and contains between 238 and 274 incidents, averaging 240 incidents per chain. As there is no public information about how many responders are operating in Seattle, we choose the typical number of responders to be 25 (roughly consistent with the ratio of responders to depots from Nashville) and perform experiments with 23, 25, and 27 responders. Finally, we train the high-level RL agent with 25 $\pm$ 3 responders.

\paragraph{Response Times}

\begin{figure*}[!ht]
\centering
\begin{subfigure}{0.3\textwidth}
\pgfplotstableread[col sep=comma,]{data_seattle/city_level_summary_5_23.csv}\regionfive
\pgfplotstableread[col sep=comma,]{data_seattle/city_level_summary_6_23.csv}\regionsix
\pgfplotstableread[col sep=comma,]{data_seattle/city_level_summary_7_23.csv}\regionseven
\begin{tikzpicture}
\begin{axis}[
      boxplot/draw direction=y,
            width=\columnwidth,
      xtick={1,2,3},
      xticklabels={{5}, {6}, {7}},
      height = 4.0cm,
      ymajorgrids,
      major grid style={draw=gray!25},
      bugsResolvedStyle/.style={},
      ylabel={Average Response Time [s]},
      xlabel={Number of Regions},
      font=\small,
    ]

\addplot+[boxplot={box extend=0.05, draw position=1}, ColorOur, solid, lshift4, fill=ColorOur!20, mark=x] table [col sep=comma, y=drl] {\regionfive};
\addplot+[boxplot={box extend=0.05, draw position=1}, ColorMCTS, solid, lshift3, fill=ColorMCTS!20, mark=x] table [col sep=comma, y=mcts] {\regionfive};
\addplot+[boxplot={box extend=0.05, draw position=1}, ColorPMedian, solid, fill=ColorPMedian!20, mark=x] table [col sep=comma, y=pmedian10] {\regionfive};
\addplot+[boxplot={box extend=0.05, draw position=1}, ColorDynamic, solid, rshift3, fill=ColorDynamic!20, mark=x] table [col sep=comma, y=greedy] {\regionfive};
\addplot+[boxplot={box extend=0.05, draw position=1}, ColorStatic, solid, rshift4, fill=ColorStatic!20, mark=x] table [col sep=comma, y=static] {\regionfive};

\addplot+[boxplot={box extend=0.05, draw position=2}, ColorOur, solid, lshift4, fill=ColorOur!20, mark=x] table [col sep=comma, y=drl] {\regionsix};
\addplot+[boxplot={box extend=0.05, draw position=2}, ColorMCTS, solid, lshift3, fill=ColorMCTS!20, mark=x] table [col sep=comma, y=mcts] {\regionsix};
\addplot+[boxplot={box extend=0.05, draw position=2}, ColorPMedian, solid, fill=ColorPMedian!20, mark=x] table [col sep=comma, y=pmedian10] {\regionsix};
\addplot+[boxplot={box extend=0.05, draw position=2}, ColorDynamic, solid, rshift3, fill=ColorDynamic!20, mark=x] table [col sep=comma, y=greedy] {\regionsix};
\addplot+[boxplot={box extend=0.05, draw position=2}, ColorStatic, solid, rshift4, fill=ColorStatic!20, mark=x] table [col sep=comma, y=static] {\regionsix};

\addplot+[boxplot={box extend=0.05, draw position=3}, ColorOur, solid, lshift4, fill=ColorOur!20, mark=x] table [col sep=comma, y=drl] {\regionseven};
\addplot+[boxplot={box extend=0.05, draw position=3}, ColorMCTS, solid, lshift3, fill=ColorMCTS!20, mark=x] table [col sep=comma, y=mcts] {\regionseven};
\addplot+[boxplot={box extend=0.05, draw position=3}, ColorPMedian, solid, fill=ColorPMedian!20, mark=x] table [col sep=comma, y=pmedian10] {\regionseven};
\addplot+[boxplot={box extend=0.05, draw position=3}, ColorDynamic, solid, rshift3, fill=ColorDynamic!20, mark=x] table [col sep=comma, y=greedy] {\regionseven};
\addplot+[boxplot={box extend=0.05, draw position=3}, ColorStatic, solid, rshift4, fill=ColorStatic!20, mark=x] table [col sep=comma, y=static] {\regionseven};
\end{axis}
\end{tikzpicture}
\caption{23 Responders
}
\end{subfigure}\hfill
\begin{subfigure}{0.3\textwidth}
\pgfplotstableread[col sep=comma,]{data_seattle/city_level_summary_5_25.csv}\regionfive
\pgfplotstableread[col sep=comma,]{data_seattle/city_level_summary_6_25.csv}\regionsix
\pgfplotstableread[col sep=comma,]{data_seattle/city_level_summary_7_25.csv}\regionseven

\begin{tikzpicture}
\begin{axis}[
      boxplot/draw direction=y,
            width=\columnwidth,
      xtick={1,2,3},
      xticklabels={{5}, {6}, {7}},
      height = 4.0cm,
      ymajorgrids,
      major grid style={draw=gray!25},
      bugsResolvedStyle/.style={},
      ylabel={Average Response Time [s]},
      xlabel={Number of Regions},
      font=\small,
    ]

\addplot+[boxplot={box extend=0.05, draw position=1}, ColorOur, solid, lshift4, fill=ColorOur!20, mark=x] table [col sep=comma, y=drl] {\regionfive};
\addplot+[boxplot={box extend=0.05, draw position=1}, ColorMCTS, solid, lshift3, fill=ColorMCTS!20, mark=x] table [col sep=comma, y=mcts] {\regionfive};
\addplot+[boxplot={box extend=0.05, draw position=1}, ColorPMedian, solid, fill=ColorPMedian!20, mark=x] table [col sep=comma, y=pmedian10] {\regionfive};
\addplot+[boxplot={box extend=0.05, draw position=1}, ColorDynamic, solid, rshift3, fill=ColorDynamic!20, mark=x] table [col sep=comma, y=greedy] {\regionfive};
\addplot+[boxplot={box extend=0.05, draw position=1}, ColorStatic, solid, rshift4, fill=ColorStatic!20, mark=x] table [col sep=comma, y=static] {\regionfive};

\addplot+[boxplot={box extend=0.05, draw position=2}, ColorOur, solid, lshift4, fill=ColorOur!20, mark=x] table [col sep=comma, y=drl] {\regionsix};
\addplot+[boxplot={box extend=0.05, draw position=2}, ColorMCTS, solid, lshift3, fill=ColorMCTS!20, mark=x] table [col sep=comma, y=mcts] {\regionsix};
\addplot+[boxplot={box extend=0.05, draw position=2}, ColorPMedian, solid, fill=ColorPMedian!20, mark=x] table [col sep=comma, y=pmedian10] {\regionsix};
\addplot+[boxplot={box extend=0.05, draw position=2}, ColorDynamic, solid, rshift3, fill=ColorDynamic!20, mark=x] table [col sep=comma, y=greedy] {\regionsix};
\addplot+[boxplot={box extend=0.05, draw position=2}, ColorStatic, solid, rshift4, fill=ColorStatic!20, mark=x] table [col sep=comma, y=static] {\regionsix};

\addplot+[boxplot={box extend=0.05, draw position=3}, ColorOur, solid, lshift4, fill=ColorOur!20, mark=x] table [col sep=comma, y=drl] {\regionseven};
\addplot+[boxplot={box extend=0.05, draw position=3}, ColorMCTS, solid, lshift3, fill=ColorMCTS!20, mark=x] table [col sep=comma, y=mcts] {\regionseven};
\addplot+[boxplot={box extend=0.05, draw position=3}, ColorPMedian, solid, fill=ColorPMedian!20, mark=x] table [col sep=comma, y=pmedian10] {\regionseven};
\addplot+[boxplot={box extend=0.05, draw position=3}, ColorDynamic, solid, rshift3, fill=ColorDynamic!20, mark=x] table [col sep=comma, y=greedy] {\regionseven};
\addplot+[boxplot={box extend=0.05, draw position=3}, ColorStatic, solid, rshift4, fill=ColorStatic!20, mark=x] table [col sep=comma, y=static] {\regionseven};

\end{axis}
\end{tikzpicture}
\caption{25 Responders}
\end{subfigure}\hfill
\begin{subfigure}{0.3\textwidth}
\pgfplotstableread[col sep=comma,]{data_seattle/city_level_summary_5_27.csv}\regionfive
\pgfplotstableread[col sep=comma,]{data_seattle/city_level_summary_6_27.csv}\regionsix
\pgfplotstableread[col sep=comma,]{data_seattle/city_level_summary_7_27.csv}\regionseven

\begin{tikzpicture}
\begin{axis}[
      boxplot/draw direction=y,
            width=\columnwidth,
      xtick={1,2,3},
      xticklabels={{5}, {6}, {7}},
      height = 4.0cm,
      ymajorgrids,
      major grid style={draw=gray!25},
      bugsResolvedStyle/.style={},
      ylabel={Average Response Time [s]},
      xlabel={Number of Regions},
      font=\small,
    ]

\addplot+[boxplot={box extend=0.05, draw position=1}, ColorOur, solid, lshift4, fill=ColorOur!20, mark=x] table [col sep=comma, y=drl] {\regionfive};
\addplot+[boxplot={box extend=0.05, draw position=1}, ColorMCTS, solid, lshift3, fill=ColorMCTS!20, mark=x] table [col sep=comma, y=mcts] {\regionfive};
\addplot+[boxplot={box extend=0.05, draw position=1}, ColorPMedian, solid, fill=ColorPMedian!20, mark=x] table [col sep=comma, y=pmedian10] {\regionfive};
\addplot+[boxplot={box extend=0.05, draw position=1}, ColorDynamic, solid, rshift3, fill=ColorDynamic!20, mark=x] table [col sep=comma, y=greedy] {\regionfive};
\addplot+[boxplot={box extend=0.05, draw position=1}, ColorStatic, solid, rshift4, fill=ColorStatic!20, mark=x] table [col sep=comma, y=static] {\regionfive};

\addplot+[boxplot={box extend=0.05, draw position=2}, ColorOur, solid, lshift4, fill=ColorOur!20, mark=x] table [col sep=comma, y=drl] {\regionsix};
\addplot+[boxplot={box extend=0.05, draw position=2}, ColorMCTS, solid, lshift3, fill=ColorMCTS!20, mark=x] table [col sep=comma, y=mcts] {\regionsix};
\addplot+[boxplot={box extend=0.05, draw position=2}, ColorPMedian, solid, fill=ColorPMedian!20, mark=x] table [col sep=comma, y=pmedian10] {\regionsix};
\addplot+[boxplot={box extend=0.05, draw position=2}, ColorDynamic, solid, rshift3, fill=ColorDynamic!20, mark=x] table [col sep=comma, y=greedy] {\regionsix};
\addplot+[boxplot={box extend=0.05, draw position=2}, ColorStatic, solid, rshift4, fill=ColorStatic!20, mark=x] table [col sep=comma, y=static] {\regionsix};

\addplot+[boxplot={box extend=0.05, draw position=3}, ColorOur, solid, lshift4, fill=ColorOur!20, mark=x] table [col sep=comma, y=drl] {\regionseven};
\addplot+[boxplot={box extend=0.05, draw position=3}, ColorMCTS, solid, lshift3, fill=ColorMCTS!20, mark=x] table [col sep=comma, y=mcts] {\regionseven};
\addplot+[boxplot={box extend=0.05, draw position=3}, ColorPMedian, solid, fill=ColorPMedian!20, mark=x] table [col sep=comma, y=pmedian10] {\regionseven};
\addplot+[boxplot={box extend=0.05, draw position=3}, ColorDynamic, solid, rshift3, fill=ColorDynamic!20, mark=x] table [col sep=comma, y=greedy] {\regionseven};
\addplot+[boxplot={box extend=0.05, draw position=3}, ColorStatic, solid, rshift4, fill=ColorStatic!20, mark=x] table [col sep=comma, y=static] {\regionseven};

\end{axis}
\end{tikzpicture}
\caption{27 Responders}
\end{subfigure}
\caption{Distribution of average response times (lower is better) with our approach (\textcolor{ColorLegendOur}{$\blacksquare$}), MCTS (\textcolor{ColorLegendMCTS}{$\blacksquare$}), $p$-median with $\alpha$ = 1.0 (\textcolor{ColorLegendPMedian}{$\blacksquare$}), greedy policy (\textcolor{ColorLegendDynamic}{$\blacksquare$}), and static policy, i.e., no proactive repositioning (\textcolor{ColorLegendStatic}{$\blacksquare$}) for 10 different sample incident chains (\textbf{Seattle}).
}
\label{fig:city_level_response_times_twenties_seatle}
\end{figure*}
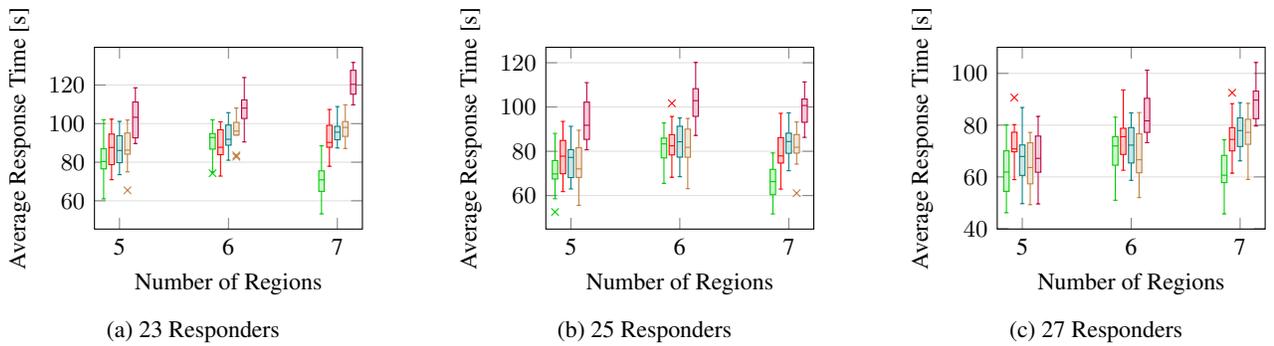

\cref{fig:city_level_response_times_twenties_seatle} shows the average response times for our DDPG-based approach compared against baseline approaches (MCTS, $p$-median-based policy, greedy policy, and static policy), based on 10 incident chains used as the evaluation set. On average, the proposed approach outperforms MCTS by 10 seconds in all cases. We observe that $p$-median, greedy, and static policy always perform poorly compared to our DDPG-based approach, and they  perform mostly poorly compared to the state-of-the-art approach (except for 3 out of 9 scenarios based on the number of regions and number of responders). \cref{fig:city_level_response_times_twenties_full_seatle} shows the average response times for our approach, MCTS, and DRLSN based on the sample chains used in the previous analysis (again, we show results with DRLSN separately as both  other approaches outperform it by a large margin, thereby making the difference between DDPG and MCTS difficult to observe when shown together). We observe that DRLSN always has response times higher than 150-200 seconds, which is at least 50\% worse than our~approach.

\begin{figure*}[!ht]
\centering
\begin{subfigure}{0.3\textwidth}
\pgfplotstableread[col sep=comma,]{data_seattle/city_level_summary_5_23.csv}\regionfive
\pgfplotstableread[col sep=comma,]{data_seattle/city_level_summary_6_23.csv}\regionsix
\pgfplotstableread[col sep=comma,]{data_seattle/city_level_summary_7_23.csv}\regionseven
\begin{tikzpicture}
\begin{axis}[
      boxplot/draw direction=y,
      width=\columnwidth,
      xtick={1,2,3},
      xticklabels={{5}, {6}, {7}},
      height = 4.0cm,
      ymajorgrids,
      major grid style={draw=gray!25},
      bugsResolvedStyle/.style={},
      ylabel={Average Response Time [s]},
      xlabel={Number of Regions},
      font=\small,
    ]

\addplot+[boxplot={box extend=0.08, draw position=1}, ColorVPG, solid, rshift, fill=ColorVPG!20, mark=x] table [col sep=comma, y=vpg] {\regionfive};
\addplot+[boxplot={box extend=0.08, draw position=1}, ColorMCTS, solid, fill=ColorMCTS!20, mark=x] table [col sep=comma, y=mcts] {\regionfive};
\addplot+[boxplot={box extend=0.08, draw position=1}, ColorOur, solid, lshift, fill=ColorOur!20, mark=x] table [col sep=comma, y=drl] {\regionfive};

\addplot+[boxplot={box extend=0.08, draw position=2}, ColorVPG, solid, rshift, fill=ColorVPG!20, mark=x] table [col sep=comma, y=vpg] {\regionsix};
\addplot+[boxplot={box extend=0.08, draw position=2}, ColorMCTS, solid, fill=ColorMCTS!20, mark=x] table [col sep=comma, y=mcts] {\regionsix};
\addplot+[boxplot={box extend=0.08, draw position=2}, ColorOur, solid, lshift, fill=ColorOur!20, mark=x] table [col sep=comma, y=drl] {\regionsix};

\addplot+[boxplot={box extend=0.08, draw position=3}, ColorVPG, solid, rshift, fill=ColorVPG!20, mark=x] table [col sep=comma, y=vpg] {\regionseven};
\addplot+[boxplot={box extend=0.08, draw position=3}, ColorMCTS, solid, fill=ColorMCTS!20, mark=x] table [col sep=comma, y=mcts] {\regionseven};
\addplot+[boxplot={box extend=0.08, draw position=3}, ColorOur, solid, lshift, fill=ColorOur!20, mark=x] table [col sep=comma, y=drl] {\regionseven};

\end{axis}
\end{tikzpicture}
\caption{23 Responders
}
\end{subfigure}\hfill
\begin{subfigure}{0.3\textwidth}
\pgfplotstableread[col sep=comma,]{data_seattle/city_level_summary_5_25.csv}\regionfive
\pgfplotstableread[col sep=comma,]{data_seattle/city_level_summary_6_25.csv}\regionsix
\pgfplotstableread[col sep=comma,]{data_seattle/city_level_summary_7_25.csv}\regionseven

\begin{tikzpicture}
\begin{axis}[
      boxplot/draw direction=y,
            width=\columnwidth,
      xtick={1,2,3},
      xticklabels={{5}, {6}, {7}},
      height = 4.0cm,
      ymajorgrids,
      major grid style={draw=gray!25},
      bugsResolvedStyle/.style={},
      ylabel={Average Response Time [s]},
      xlabel={Number of Regions},
      font=\small,
    ]

\addplot+[boxplot={box extend=0.08, draw position=1}, ColorVPG, solid, rshift, fill=ColorVPG!20, mark=x] table [col sep=comma, y=vpg] {\regionfive};
\addplot+[boxplot={box extend=0.08, draw position=1}, ColorMCTS, solid, fill=ColorMCTS!20, mark=x] table [col sep=comma, y=mcts] {\regionfive};
\addplot+[boxplot={box extend=0.08, draw position=1}, ColorOur, solid, lshift, fill=ColorOur!20, mark=x] table [col sep=comma, y=drl] {\regionfive};

\addplot+[boxplot={box extend=0.08, draw position=2}, ColorVPG, solid, rshift, fill=ColorVPG!20, mark=x] table [col sep=comma, y=vpg] {\regionsix};
\addplot+[boxplot={box extend=0.08, draw position=2}, ColorMCTS, solid, fill=ColorMCTS!20, mark=x] table [col sep=comma, y=mcts] {\regionsix};
\addplot+[boxplot={box extend=0.08, draw position=2}, ColorOur, solid, lshift, fill=ColorOur!20, mark=x] table [col sep=comma, y=drl] {\regionsix};

\addplot+[boxplot={box extend=0.08, draw position=3}, ColorVPG, solid, rshift, fill=ColorVPG!20, mark=x] table [col sep=comma, y=vpg] {\regionseven};
\addplot+[boxplot={box extend=0.08, draw position=3}, ColorMCTS, solid, fill=ColorMCTS!20, mark=x] table [col sep=comma, y=mcts] {\regionseven};
\addplot+[boxplot={box extend=0.08, draw position=3}, ColorOur, solid, lshift, fill=ColorOur!20, mark=x] table [col sep=comma, y=drl] {\regionseven};

\end{axis}
\end{tikzpicture}
\caption{25 Responders}
\end{subfigure}\hfill
\begin{subfigure}{0.3\textwidth}
\pgfplotstableread[col sep=comma,]{data_seattle/city_level_summary_5_27.csv}\regionfive
\pgfplotstableread[col sep=comma,]{data_seattle/city_level_summary_6_27.csv}\regionsix
\pgfplotstableread[col sep=comma,]{data_seattle/city_level_summary_7_27.csv}\regionseven

\begin{tikzpicture}
\begin{axis}[
      boxplot/draw direction=y,
            width=\columnwidth,
      xtick={1,2,3},
      xticklabels={{5}, {6}, {7}},
      height = 4.0cm,
      ymajorgrids,
      major grid style={draw=gray!25},
      bugsResolvedStyle/.style={},
      ylabel={Average Response Time [s]},
      xlabel={Number of Regions},
      font=\small,
    ]

\addplot+[boxplot={box extend=0.08, draw position=1}, ColorVPG, solid, rshift, fill=ColorVPG!20, mark=x] table [col sep=comma, y=vpg] {\regionfive};
\addplot+[boxplot={box extend=0.08, draw position=1}, ColorMCTS, solid, fill=ColorMCTS!20, mark=x] table [col sep=comma, y=mcts] {\regionfive};
\addplot+[boxplot={box extend=0.08, draw position=1}, ColorOur, solid, lshift, fill=ColorOur!20, mark=x] table [col sep=comma, y=drl] {\regionfive};

\addplot+[boxplot={box extend=0.08, draw position=2}, ColorVPG, solid, rshift, fill=ColorVPG!20, mark=x] table [col sep=comma, y=vpg] {\regionsix};
\addplot+[boxplot={box extend=0.08, draw position=2}, ColorMCTS, solid, fill=ColorMCTS!20, mark=x] table [col sep=comma, y=mcts] {\regionsix};
\addplot+[boxplot={box extend=0.08, draw position=2}, ColorOur, solid, lshift, fill=ColorOur!20, mark=x] table [col sep=comma, y=drl] {\regionsix};

\addplot+[boxplot={box extend=0.08, draw position=3}, ColorVPG, solid, rshift, fill=ColorVPG!20, mark=x] table [col sep=comma, y=vpg] {\regionseven};
\addplot+[boxplot={box extend=0.08, draw position=3}, ColorMCTS, solid, fill=ColorMCTS!20, mark=x] table [col sep=comma, y=mcts] {\regionseven};
\addplot+[boxplot={box extend=0.08, draw position=3}, ColorOur, solid, lshift, fill=ColorOur!20, mark=x] table [col sep=comma, y=drl] {\regionseven};

\end{axis}
\end{tikzpicture}
\caption{27 Responders}
\end{subfigure}
\caption{Distribution of average response times (lower is better) with our approach (\textcolor{ColorLegendOur}{$\blacksquare$}), MCTS (\textcolor{ColorLegendMCTS}{$\blacksquare$}), and DRLSN (\textcolor{ColorLegendVPG}{$\blacksquare$}) for 10 different sample incident chains \textbf{(Seattle)}. In this figure, we plot the same data for our approach and MCTS as in \cref{fig:city_level_response_times_twenties_seatle}; the only difference is the inclusion of DRLSN, which changes the scaling of the vertical axis.
}
\label{fig:city_level_response_times_twenties_full_seatle}
\end{figure*}
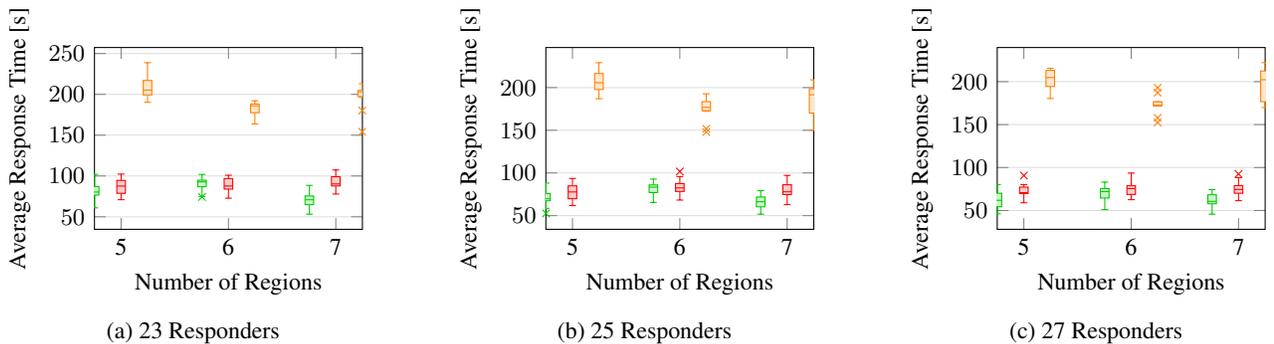

\paragraph{Learning Curves}

\pgfplotstableread[col sep=comma,]{data_seattle/seattle_7_regions_region_0.csv}\regionsevenregionzero

\pgfplotstableread[col sep=comma,]{data_seattle/seattle_7_regions_region_1.csv}\regionsevenregionone

\pgfplotstableread[col sep=comma,]{data_seattle/seattle_7_regions_region_2.csv}\regionsevenregiontwo

\pgfplotstableread[col sep=comma,]{data_seattle/seattle_7_regions_city.csv}\regionsevenregioncity

\begin{figure*}[th]
\begin{subfigure}{0.32\textwidth}
\begin{tikzpicture}
\begin{axis}[
      width=\linewidth,
      height = 5cm,
      bugsResolvedStyle/.style={},
      ylabel={Response Time [Seconds]},
      xlabel={Training Episodes},
    ]

\addplot[very thick, ColorRandom] table [x=index, y=random_mean, col sep=comma] {\regionsevenregionzero};

\addplot[very thick, ColorRandom!50, name path=A] table [x=index, y=random_min, col sep=comma] {\regionsevenregionzero};
\addplot[very thick, ColorRandom!50, name path=B] table [x=index, y=random_max, col sep=comma] {\regionsevenregionzero};

\addplot[fill=ColorRandom, opacity=0.25] fill between[of=A and B];

\addplot[very thick, ColorOur] table [x=index, y=response_time_mean, col sep=comma] {\regionsevenregionzero};

\addplot[very thick, ColorOur!50, name path=A] table [x=index, y=response_time_min, col sep=comma] {\regionsevenregionzero};
\addplot[very thick, ColorOur!50, name path=B] table [x=index, y=response_time_max, col sep=comma] {\regionsevenregionzero};

\addplot[fill=ColorOur, opacity=0.25] fill between[of=A and B];
\end{axis}
\end{tikzpicture}
\caption{Region \#1}
\label{fig:seattle_region_1_learning_curve}
\end{subfigure}%
\begin{subfigure}{0.32\textwidth}
\begin{tikzpicture}
\begin{axis}[
      width=\linewidth,
      height = 5cm,
      bugsResolvedStyle/.style={},
      ylabel={Response Time [Seconds]},
      xlabel={Training Episodes},
    ]

\addplot[very thick, ColorRandom] table [x=index, y=random_mean, col sep=comma] {\regionsevenregionone};

\addplot[very thick, ColorRandom!50, name path=A] table [x=index, y=random_min, col sep=comma] {\regionsevenregionone};
\addplot[very thick, ColorRandom!50, name path=B] table [x=index, y=random_max, col sep=comma] {\regionsevenregionone};

\addplot[fill=ColorRandom, opacity=0.25] fill between[of=A and B];

\addplot[very thick, ColorOur] table [x=index, y=response_time_mean, col sep=comma] {\regionsevenregionone};

\addplot[very thick, ColorOur!50, name path=A] table [x=index, y=response_time_min, col sep=comma] {\regionsevenregionone};
\addplot[very thick, ColorOur!50, name path=B] table [x=index, y=response_time_max, col sep=comma] {\regionsevenregionone};

\addplot[fill=ColorOur, opacity=0.25] fill between[of=A and B];
\end{axis}
\end{tikzpicture}
\caption{Region \#2}
\label{fig:seattle_region_2_learning_curve}
\end{subfigure}%
\begin{subfigure}{0.32\textwidth}
\begin{tikzpicture}
\begin{axis}[
      width=\linewidth,
      height = 5cm,
      bugsResolvedStyle/.style={},
      ylabel={Response Time [Seconds]},
      xlabel={Training Episodes},
    ]

\addplot[very thick, ColorRandom] table [x=index, y=random_mean, col sep=comma] {\regionsevenregiontwo};

\addplot[very thick, ColorRandom!50, name path=A] table [x=index, y=random_min, col sep=comma] {\regionsevenregiontwo};
\addplot[very thick, ColorRandom!50, name path=B] table [x=index, y=random_max, col sep=comma] {\regionsevenregiontwo};

\addplot[fill=ColorRandom, opacity=0.25] fill between[of=A and B];

\addplot[very thick, ColorOur] table [x=index, y=response_time_mean, col sep=comma] {\regionsevenregiontwo};

\addplot[very thick, ColorOur!50, name path=A] table [x=index, y=response_time_min, col sep=comma] {\regionsevenregiontwo};
\addplot[very thick, ColorOur!50, name path=B] table [x=index, y=response_time_max, col sep=comma] {\regionsevenregiontwo};

\addplot[fill=ColorOur, opacity=0.25] fill between[of=A and B];
\end{axis}
\end{tikzpicture}
\caption{Region \#3}
\label{fig:seattle_region_3_learning_curve}
\end{subfigure}
\caption{Evolution of the performance of the low-level policy $\mu_{\OneRegion}$ throughout the training process for regions 1, 2, and 3 of the 7-region decomposition in \textbf{Seattle}, measured as the average response time. The dark green line (\textcolor{ColorLegendOurThick}{$\blacksquare$})  indicates the average of 10 different policies (trained on the given number of episodes), which are evaluated on 5 different sample chains with the number of responders ranging from 1 to $|\RegionDepots|$. The light green area (\textcolor{ColorLegendOur}{$\blacksquare$}) indicates the 10th to 90th percentiles of average response times  over 10 different policies. The dark gray line (\textcolor{ColorLegendRandomThick}{$\blacksquare$})  indicates the mean of average response time over the same set of samples when using a random policy, and the light gray area (\textcolor{ColorLegendRandom}{$\blacksquare$}) indicates the 10th to 90th percentiles of average response times over 15 different random policies.}
\end{figure*}

\begin{figure}[t]
\begin{tikzpicture}
\begin{axis}[
      width=\linewidth,
      height = 5cm,
      bugsResolvedStyle/.style={},
      ylabel={Response Time [Seconds]},
      xlabel={Training Episodes},
    ]
\addplot[very thick, ColorMCTS] table [x=index, y=mcts, col sep=comma] {\regionsevenregioncity};

\addplot[very thick, ColorOur] table [x=index, y=response_time_mean, col sep=comma] {\regionsevenregioncity};

\addplot[very thick, ColorOur!50, name path=A] table [x=index, y=response_time_min, col sep=comma] {\regionsevenregioncity};
\addplot[very thick, ColorOur!50, name path=B] table [x=index, y=response_time_max, col sep=comma] {\regionsevenregioncity};

\addplot[fill=ColorOur, opacity=0.25] fill between[of=A and B];
\end{axis}
\end{tikzpicture}
\caption{Evolution of the performance of the policy $\mu_{h}$ throughout the training process for the high-level decision agent of the 7-region decomposition in \textbf{Seattle}, measured as the average response time. The dark green line (\textcolor{ColorLegendOurThick}{$\blacksquare$}) indicates the average of 10 different policies (trained on the given number of episodes), which are evaluated on 5 different sample chains with the number of responders in the entire city ranging from 23 to 27. The light green area (\textcolor{ColorLegendOur}{$\blacksquare$}) indicates the 10th to 90th percentiles of average response times over 5 different policies. The red line (\textcolor{ColorLegendMCTSThick}{$\blacksquare$}) indicates the average response times obtained when using the state-of-the-art MCTS approach on the same 5 sample chains with the number of responders in the entire city ranging from 23 to 27.}
\label{fig:seattle_city_learning_curve}
\end{figure}
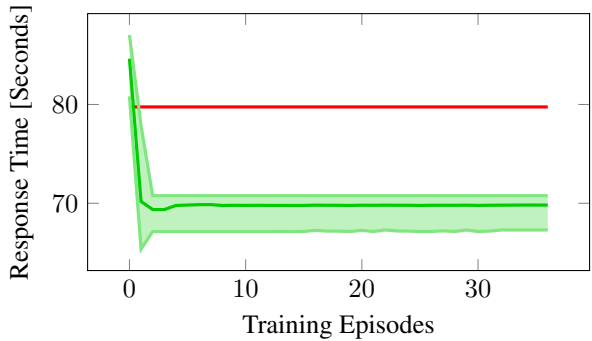

\cref{fig:seattle_region_1_learning_curve,fig:seattle_region_2_learning_curve,fig:seattle_region_3_learning_curve} show the performance of the low-level policy $\mu_{g}$ for regions 1, 2, and 3 of the 7-region decomposition in Seattle.
The vertical axis indicates the average response times, and the horizontal axis indicates the number of training episodes. The light green area (\textcolor{ColorLegendOur}{$\blacksquare$}) represents the 10th to  90th percentiles of average response times over 10 different policies. The light gray area (\textcolor{ColorLegendRandom}{$\blacksquare$}) represents the 10th to 90th percentiles of the average response times over 15 different random policies. 
We observe that our trained policy is significantly better than the random policy and learns fast.

\cref{fig:seattle_city_learning_curve} shows the performance of the policy $\mu_{h}$ throughout the training process for  the high-level decision agent of the 7-region decomposition in Seattle, measured as the average response time. The light green area  represents the 10th to 90th percentiles of the average response times over 10 different policies. The red line indicates the average response times obtained using the state-of-the-art MCTS approach on the same 5 sample chains with the number of responders ranging from 23 to 27. We observe that our high-level policy training helps to reduce the response time compared to the state-of-the-art approach.

\subsection{Architecture Search for Low-Level RL Agent}
\label{app:architecture_search}

We experiment with architectures such as LSTM, TrXL \cite{vaswani2017attention,dai2019transformer,parisotto2020stabilizing}, and Gated Transformer-XL  (GTrXL) \cite{parisotto2020stabilizing,parisotto2021efficient} as neural-network architecture choices for the actor in the low-level agent. We perform a random architecture search to obtain the best hyperparameters for each architecture (i.e., LSTM, TrXL, GTrXL). Accordingly, we tune the following hyperparameters in the actor-network using TrXL:
\begin{itemize}[noitemsep, topsep=0pt]
\item Hidden layers in MLP: 1 / 2 / 3 layers
\item Number of neurons per hidden layers in MLP: 32 / 64 / 128 / 256 neurons per layer
\item Dropout rate in MLP (after every hidden layer): 0.0 / 0.0125 / 0.025 / 0.05 / 0.1
\item Number of attention heads:  1 / 2 / 3 / 4 / 5 heads
\item Number of layers ($N$): 1 / 2 / 3 layers
\end{itemize}

\begin{table*}
\centering
\caption{Best Hyperparameters for Low-Level Agent Actor using TrXL (\textbf{Nashville})}
\label{tab:best_architectures_nashville}

\begin{tabular}{|c|c|c|c|c|c|}\hline
\textbf{Number of  regions} & \multicolumn{5}{c|}{\textbf{5}} \\ \hline
Region identifier & 0 & 1 & 2 & 3 & 4\\ \hline\hline
Number of  layers ($N$)& 1 & 3 & 3 & 2 & 3\\ \hline
Number of  heads in MHA & 5 & 3 & 3 & 1 & 5\\ \hline
MLP& 256, 128 & 128, 64 & 128, 64 & 256, 128 & 64\\ \hline
Dropout rate & 0.1 & 0.1 & 0.1 & 0.0 & 0.05\\ \hline
\end{tabular}

\vspace{1em}

\begin{tabular}{|c|c|c|c|c|c|c|}\hline
\textbf{Number of  regions} & \multicolumn{6}{c|}{\textbf{6}} \\ \hline
Region identifier & 0 & 1 & 2 & 3 & 4 & 5\\ \hline\hline
Number of  layers ($N$)& 2 & 3 & 3 & 3 & 1 & 2\\ \hline
Number of  heads in MHA & 1 & 2 & 5 & 5 & 2 & 4\\ \hline
MLP & 256, 128 & 64 & 32 & 64 & 128, 64 & 256, 128\\ \hline
Dropout rate & 0.05 & 0.0 & 0.1 & 0.1 & 0.0 & 0.0\\ \hline
\end{tabular}

\vspace{1em}

\begin{tabular}{|c|c|c|c|c|c|c|c|}\hline
\textbf{Number of  regions} & \multicolumn{7}{c|}{\textbf{7}} \\ \hline
Region identifier & 0 & 1 & 2 & 3 & 4 & 5 & 6\\ \hline\hline
Number of  layers ($N$)& 2 & 3 & 1 & 1 & 2 & 2 & 1\\ \hline
Number of  heads in MHA & 3 & 1 & 3 & 5 & 4 & 3 & 5\\ \hline
MLP & 256, 128, 64 & 256, 128, 64 & 128, 64 & 64 & 32 & 256, 128 & 64\\ \hline
Dropout rate & 0.1 & 0.1 & 0.1 & 0.05 & 0.0 & 0.05 & 0.0\\ \hline
\end{tabular}

\end{table*}

\begin{table*}
\centering
\caption{Best Hyperparameters for Low-Level Agent Actor using TrXL (\textbf{Seattle})}
\label{tab:best_architectures_seattle}

\begin{tabular}{|c|c|c|c|c|c|}\hline
\textbf{Number of regions} & \multicolumn{5}{c|}{\textbf{5}} \\ \hline
Region identifier & 0 & 1 & 2 & 3 & 4\\ \hline\hline
Number of layers ($N$)& 3 & 3 & 1 & 3 & 2\\ \hline
Number of heads in MHA & 1 & 3 & 3 & 1 & 5\\ \hline
MLP & 256 & 256 & 32 & 256 & 32\\ \hline
Dropout rate & 0.0125 & 0.0 & 0.1 & 0.0125 & 0.0125\\ \hline
\end{tabular}

\vspace{1em}

\begin{tabular}{|c|c|c|c|c|c|c|}\hline
\textbf{Number of regions} & \multicolumn{6}{c|}{\textbf{6}} \\ \hline
Region identifier & 0 & 1 & 2 & 3 & 4 & 5\\ \hline\hline
Number of layers ($N$)& 1 & 1 & 1 & 1 & 2 & 1\\ \hline
Number of heads in MHA & 5 & 3 & 5 & 5 & 4 & 5\\ \hline
MLP & 128 & 128, 64 & 256 & 128, 64 & 256 & 128\\ \hline
Dropout rate & 0.1 & 0.0125 & 0.1 & 0.025 & 0.0125 & 0.1\\ \hline
\end{tabular}

\vspace{1em}

\begin{tabular}{|c|c|c|c|c|c|c|c|}\hline
\textbf{Number of regions} & \multicolumn{7}{c|}{\textbf{7}} \\ \hline
Region identifier & 0 & 1 & 2 & 3 & 4 & 5 & 6\\ \hline\hline
Number of layers ($N$)& 1 & 3 & 1 & 2 & 2 & 2 & 2\\ \hline
Number of heads in MHA & 1 & 1 & 2 & 5 & 4 & 3 & 2\\ \hline
MLP & 64 & 256 & 64 & 256, 128, 64 & 64 & 128, 64 & 256, 128, 64\\ \hline
Dropout rate & 0.025 & 0.1 & 0.05 & 0.0 & 0.0 & 0.05 & 0.0125\\ \hline
\end{tabular}

\end{table*}

We perform the architecture search for each region in the segmentations (i.e., 5, 6, and 7 regions). We terminate the architecture search after obtaining an actor-network that can outperform the competitive baseline using MCTS. 
\cref{tab:best_architectures_nashville,tab:best_architectures_seattle} show the best hyperparameters for the TrXL based low-level agent for each region (with  5, 6, and 7 region segmentations) for Nashville and Seattle data, respectively.

\section{Statistical Tests of Significance}
\label{app:statistical_tests}

In this section, we present the results of statistical tests on the significance of our numerical results. %
Specifically, we present paired two-sample permutation tests comparing the average response times of our proposed approach to the average response times of each baseline approach based on a sample of 10 incident chains in each case.
We perform these tests for various numbers of regions (5, 6, and 7) and various numbers of responders for both Nashville and Seattle.
Since our goal is to compare the average response times, we establish the following null and alternative hypotheses:
\begin{itemize}[noitemsep, topsep=0pt]
    \item \textit{Null Hypothesis} ($H_0$): mean of the response times obtained using our DDPG-based approach is the same as the mean of the response times obtained using the baseline (i.e., MCTS, $p$-median-based policy, greedy policy, static policy, or DRLSN);
    \item \textit{Alternate Hypothesis} ($H_1$): mean of the response times obtained using our DDPG-based approach is significantly different from the mean of the response times obtained using the baseline (i.e., MCTS, $p$-median-based policy, greedy policy, static policy, or DRLSN).
\end{itemize}
We use the difference between the means as our test statistic.

\begin{table*}[!ht]
\centering
\caption{$p$-Values to Reject the Null Hypothesis (\textbf{Nashville})}
\label{tab:p_values_nashville}
\begin{tabular}{|c|c|c|c|c|c|c|c|c|c|}\hline
\textbf{Number of regions} & \multicolumn{3}{c|}{\textbf{5}} & \multicolumn{3}{c|}{\textbf{6}} & \multicolumn{3}{c|}{\textbf{7}} \\ \hline
\textbf{Number of responders} & 24 & 26 & 28 & 24 & 26 & 28 & 24 & 26 & 28\\ \hline\hline
MCTS & 0.01 & 0.01 & 0.02 & \textbf{0.14} & \textbf{0.20} & 0.02 & 0.00 & 0.00 & 0.00 \\ \hline
$p$-median-based policy & 0.00 & 0.00 & 0.00 & 0.00 & 0.00 & 0.00 & 0.00 & 0.00 & 0.00 \\ \hline
greedy policy & 0.00 & 0.00 & 0.00 & 0.00 & 0.00 & 0.00 & 0.00 & 0.00 & 0.00 \\ \hline
static policy & 0.00 & 0.00 & 0.00 & 0.00 & 0.00 & 0.00 & 0.00 & 0.00 & 0.00 \\ \hline
DRLSN & 0.00 & 0.00 & 0.00 & 0.00 & 0.00 & 0.00 & 0.00 & 0.00 & 0.00 \\ \hline
\end{tabular}

\end{table*}

\begin{table*}[!ht]
\centering
\caption{$p$-Values to Reject the Null Hypothesis (\textbf{Seattle})}
\label{tab:p_values_seattle}

\begin{tabular}{|c|c|c|c|c|c|c|c|c|c|}\hline
\textbf{Number of regions} & \multicolumn{3}{c|}{\textbf{5}} & \multicolumn{3}{c|}{\textbf{6}} & \multicolumn{3}{c|}{\textbf{7}} \\ \hline
\textbf{Number of responders} & 23 & 25 & 27 &  23 & 25 & 27 &  23 & 25 & 27\\ \hline\hline
MCTS & 0.04 & 0.01 & 0.00 & \textbf{0.37} & \textbf{0.36} & 0.04 & 0.00 & 0.00 & 0.00 \\ \hline
$p$-median-based policy & 0.03 & 0.02 & 0.02 & \textbf{0.07} & \textbf{0.07} & \textbf{0.16} & 0.00 & 0.00 & 0.00 \\ \hline
greedy policy & \textbf{0.05} & \textbf{0.09} & 0.02 & 0.00 & \textbf{0.55} & \textbf{0.19} & 0.00 & 0.00 & 0.00 \\ \hline
static policy & 0.00 & 0.00 & 0.00 & 0.00 & 0.00 & 0.00 & 0.00 & 0.00 & 0.00 \\ \hline
DRLSN & 0.00 & 0.00 & 0.00 & 0.00 & 0.00 & 0.00 & 0.00 & 0.00 & 0.00 \\ \hline
\end{tabular}

\end{table*}

\cref{tab:p_values_nashville,tab:p_values_seattle} shows the $p$-values to reject the null hypothesis based on samples of 10 incidents chains for Nashville and Seattle, respectively. 
We observe that for Nashville, \emph{the average response times obtained using our DDPG-based approach are significantly better than those of the baselines} (considering the significance threshold for the $p$-value to be 5\%), except in two cases: 6 region decomposition with 24 and 26 responders with the state-of-the-art MCTS approach.
Note that even in these cases, the $p$-value is low, suggesting that the null hypothesis is likely false, and the key advantage of our approach over MCTS is not in lowering response time but in lowering running time by several orders of magnitude.
Similarly, for Seattle, we observe that the average response times obtained using our DDPG-based approach are significantly better than those of the baselines in most cases, except for a few scenarios.
Again, we observe that MCTS is close in two cases, but the main advantage of our approach is significantly lowering running time in all cases.
We also observe similarity in a few other cases.
Note that we performed the tests with relatively small samples (10 chains, i.e., 10 values); more extensive experiments could lead to lower $p$-values.

\section{Extended Related Work}
\label{app:extended_related_work}

In \cref{sec:related}, we provided a concise summary of the most relevant prior work due to the limited space. In this section, we provide a broader discussion of related work.

\subsection{Emergency Response Management}
\label{app:erw_ems}
Prior works use \textit{centralized} \cite{geccopaper,ji2019deep}, \textit{decentralized} \cite{pettet2020algorithmic}, and \textit{hierarchical} \cite{pettet2021hierarchical2} approaches to solve the ERM reallocation problem. \citet{geccopaper} use a genetic algorithm-based solution to the responder redistribution problem. However, \citet{geccopaper} reposition responders only at shift changes (e.g., from day shift to night shift) throughout the day. In contrast, our approach performs proactive reallocation whenever a new incident arrives or when the arrival rates of incidents change. \citet{ji2019deep} reallocate each responder once it finishes serving its current incident assignment. In contrast, in our approach, we allow complete reallocation at each incident arrival and at each change in the rate of future incidents. In \citet{ji2019deep}'s approach, during each reallocation step, the trained RL agent outputs a score for all available depots (based on features such as nearby incident rates for the depot and the expected time for the nearest-$k$ responders to reach the depot) and chooses to assign the responder to the depot with the highest score. In our approach, we consider similar features when performing the reallocation: we consider features such as nearby incident rates for each depot and the expected time for a responder to reach a depot.

\subsection{Dispatching Problem}
There are two key differences between order-dispatching approaches, such as the works \cite{zhou2019multi,li2019efficient} and proactive allocation for emergency response. 
First, order-dispatching approaches focus on responding to requests by optimizing the dispatch of vehicles when new requests arrive. In emergency-response management (ERM), dispatch decisions cannot be optimized as ERM systems are typically mandated to always dispatch the nearest responder \cite{pettet2021hierarchical2,ji2019deep,mukhopadhyay2020review}. Therefore, similar to prior work, we focus on optimizing the allocation of responders in anticipation of the arrival of future requests. Second, more importantly, both works \cite{zhou2019multi,li2019efficient} assume high-level states and actions, defined in terms of numbers of vehicles in given areas. In contrast, we consider fine-grained states and actions, defined in terms of allocating specific vehicles to specific locations (depots). High-level states and actions are appropriate for managing large-scale ride-sharing services with hundreds (or thousands) of vehicles \cite{zhou2019multi,li2019efficient}; however, emergency response requires fine-grained management of individual responders as every second counts. Our work focuses on the computational challenges that arise from the combinatorial nature of these fine-grained states and actions.

\subsection{Transformers}
\label{app:erw_trans}

\citet{dai2019transformer} introduce a transformer variant that can work with variable-length inputs. \citet{parisotto2020stabilizing} introduce stabilization over TrXL, via performing normalization before MHA and MLP; further, they concatenate the residual connections using a Gated Layer, and the complete architecture is often called Gated TrXL (GTrXL). To train our low-level RL agents, we try both TrXL and GTrXL variants. However, in contrast to the results of
\citet{parisotto2020stabilizing}, we find that TrXL can perform better than GTrXL in our problem setting. Accordingly, we use TrXL as our neural network architecture for low-level RL agents.

\subsection{Hierarchical Reinforcement Learning}
\label{app:erw_hrl}
Hierarchical Reinforcement Learning (HRL) solves complex tasks by training agents to make decisions over multiple levels of temporal abstraction \cite{pateria2021hierarchical,hutsebaut2022hierarchical,xu2023haven}. \citet{pateria2021hierarchical} categorize HRL based on the number of agents in the system (i.e., single agent or multi-agent), nature of the tasks (i.e., heterogeneous or homogeneous) and whether or not the policy has to discover the sub-tasks. \citet{xu2023haven,jin2019coride} apply HRL in MARL that operates cooperatively. In contrast to HRL, in our work, we consider a MARL system with heterogeneous agents (i.e., a set of low-level agents with the task of repositioning responders within their designated regions and a high-level agent with the task of redistributing responders among regions) that acts cooperatively to achieve the goal of minimizing the response time for serving incidents.

\section{Rationale behind Application of DDPG}
\label{app:why_ddpg}

Our approach uses the well-known DDPG algorithm to train both low-level and high-level agents. However, the DDPG actor outputs a continuous action. In contrast, our environment expects a discrete action that represents the assignment of  responders to  depots or the redistribution of responders to  regions. To discretize the continuous actor action, we use combinatorial optimization techniques.

One seemingly trivial solution to the problem  above is using an RL algorithm that works well with discrete action spaces, such as Deep Q-Learning Network (DQN) \cite{mnih2015human} or Soft-Actor Critic (SAC) \cite{haarnoja2018soft}. While these RL algorithms are often straightforward in terms of computing the state-action value, they run into scalability issues when the space of possible actions grows prohibitively large. For example, consider a region with 10 depots and 10 responders. There are  $10 ! \approx 3 \times 10^{6}$ possible ways  to reposition responders between depots. Even if we can infer the value of a single repositioning action at the speed of $0.01$ seconds, we still require $10^4$ seconds (around 3 hours) to compute the values of all possible allocations. In contrast, using DDPG can make a single decision in a fraction of a second (0.22 seconds).

Another way to tackle the discretization problem is to apply an RL approach that assumes a parametric distribution over the action space, whose parameters are to be estimated by the actor (e.g., Proximal Policy Optimization (PPO) \cite{schulman2017proximal}, SAC \cite{haarnoja2018soft}, Q-functionals \cite{lobel2023q}). In this case, the actor output would be similar to that of our DDPG actor (i.e., compact marginal of a larger distribution), losing the advantages of these RL algorithms while having to deal with additional challenges, such as sampling from a combinatorial discrete action space. %

\fi

\end{document}